\documentclass[letterpaper]{article}

\usepackage{microtype}
\usepackage{graphicx}
\usepackage{subcaption}
\usepackage{booktabs} 
\usepackage{hyperref}
\usepackage{xspace}

\usepackage[accepted]{icml2026}

\usepackage{amsmath}
\usepackage{amssymb}
\usepackage{mathtools}
\usepackage{amsthm}
\usepackage[capitalize,noabbrev]{cleveref}

\usepackage{xcolor}              
\usepackage[normalem]{ulem}      
\usepackage{todonotes}       

\theoremstyle{plain}

\theoremstyle{definition}

\theoremstyle{remark}

\newcommand\paraspace{}
\providecommand\parab[1]{\paraspace\noindent\textbf{#1}}

\newcommand{\secref}[1]{\S\ref{#1}}
\newcommand{\figref}[1]{Figure~\ref{#1}}

\newcommand{\sysname}{\textsc{Mosaic}\xspace}

\usepackage{todonotes}
\usepackage{tikz}

\usepackage[linesnumbered,ruled,vlined,algo2e]{algorithm2e}
\usepackage{xcolor}

\SetCommentSty{mycommfont}

\SetKwComment{tcp}{// }{}

\newcommand*\circlednum[1]{%
  \tikz[baseline=(char.base)]{
    \node[shape=circle,draw,inner sep=1pt, font=\scriptsize] (char) {#1};}}

\newcommand{\etc}{\emph{etc.}\xspace}

\newcommand{\eg}{\emph{e.g.,}\xspace}

\newif\ifshowtodos

\newcommand{\yitao}[1]{\ifshowtodos\todo[author=\textbf{Yitao},color=orange!25,inline]{#1}\fi}

\newif\ifshowrevise
\showrevisetrue  

\newcommand{\revise}[2]{%
  \ifshowrevise
    \ifx&#1&\else{\color[RGB]{0,0,192}\sout{#1}}\fi 
    \ifx&#1&\else\ \fi                              
    \ifx&#2&\else{\color[RGB]{192,0,0}{#2}}\fi      
  \else#2\fi                                        
}

\graphicspath{{figs/}}

\icmltitlerunning{\sysname: Unlocking Over 30$\times$ Context Length for Diffusion LLMs Inference}

\begin{document}

\twocolumn[
\icmltitle{\sysname: Unlocking Over 30$\times$ Context Length for Diffusion LLMs Inference via Global Memory Planning and Dynamic Peak Taming}

\begin{icmlauthorlist}
    \icmlauthor{Liang Zheng}{tju}
    \icmlauthor{Bowen Shi}{tju}
    \icmlauthor{Yitao Hu}{tju}
    \icmlauthor{Jiawei Zhang}{tju}
    \icmlauthor{Ruofan Li}{tju}
    \icmlauthor{Guotao Yang}{tju}
    \icmlauthor{Zhixin Zhao}{tju}
    \icmlauthor{Zhengchao Wang}{tju}
    \icmlauthor{Sheng Chen}{tju}
    \icmlauthor{Wenxin Li}{tju}
    \icmlauthor{Dezhi Ran}{pku}
    \icmlauthor{Tao Xie}{pku}
    \icmlauthor{Keqiu Li}{tju}
 
\end{icmlauthorlist}

\icmlaffiliation{tju}{Tianjin University, China}
\icmlaffiliation{pku}{Peking University, China}

\icmlcorrespondingauthor{Yitao Hu}{yitao@tju.edu.cn}

\icmlkeywords{Machine Learning, ICML}

\vskip 0.3in
]

\printAffiliationsAndNotice{}

\begin{abstract}

Diffusion-based large language models (dLLMs) have emerged as a promising alternative to autoregressive models, leveraging simultaneous denoising to enable global planning and iterative refinement. These properties make dLLMs attractive for long-context generation. However, deploying dLLMs faces a prohibitive memory barrier, as existing inference systems are inefficient for the diffusion paradigm.
We observe that current inference systems are misaligned with dLLMs. Unlike autoregressive models, whose memory footprint is dominated by the KV-Cache, dLLMs are bottlenecked by transient activations rematerialized per step. Moreover, generic memory reuse mechanisms lack the global visibility to handle dynamic memory peaks of dLLMs, which alternate between logits and feed-forward networks.
To address these challenges, we present \sysname, a memory-efficient inference system that shifts dLLM execution from local, static management to a global, dynamic paradigm. \sysname integrates (i) a mask-only logits kernel eliminating redundant activation materialization, (ii) a lazy chunking optimizer using online heuristics to tame dynamic memory peaks, and (iii) a global memory manager leveraging virtual addressing to mitigate memory fragmentation.
Evaluations show that \sysname reduces the memory peak-to-average ratio by 2.71$\times$ on average and increases the maximum inference sequence length on identical hardware by 15.30--32.34$\times$.  Crucially, \sysname is training-free and preserves exact model outputs, while reducing end-to-end latency by 2.5\%--55.4\%. Our code is publicly available at \url{https://github.com/flashserve/Mosaic}.
\end{abstract}

\section{Introduction}

\yitao{@Yitao, revise all figures in the paper, including caption, color style, paper position/location, \etc}

While autoregressive (AR) models have driven recent breakthroughs in generative AI~\cite{zhao2023survey,minaee2024large}, diffusion-based large language models (dLLMs) have emerged as a distinct paradigm. Unlike the sequential, left-to-right generation of AR models, dLLMs progressively denoise the entire sequence simultaneously~\cite{nie2025large,zhu2025llada,ye2025dream}. This capability facilitates global planning and iterative refinement, making dLLMs a promising paradigm for maintaining long-range consistency~\cite{sahoo2024simple,yu2025discrete}.

A key trend in dLLM is the shift toward long-context generation~\cite{liu2025longllada,he2025ultrallada}. This capability unlocks advanced applications such as repository-level code generation~\cite{xie2025dream} and massive-context information filling. Despite this potential, deploying long-context dLLMs remains constrained by a prohibitive memory capacity barrier stemming from system inefficiencies.

Specifically, existing memory optimizations for LLMs prioritize KV-Cache~\cite{kwon2023efficient,prabhu2025vattention}, the primary bottleneck in autoregressive generation; however, these techniques do not transfer to dLLMs. Our profiling reveals a fundamental shift: dLLMs are bottlenecked by \textit{transient activations} recomputed at every step. While general-purpose reuse mechanisms address activation memory to some extent~\cite{pisarchyk2020efficient,ansel2024pytorch}, they lack the global visibility to adapt to dLLMs' \textit{dynamic peaks}, which toggle between logits and feed-forward networks (FFNs). Consequently, applying existing methods results in significant memory fragmentation and redundancy, necessitating a system tailored to these distinct patterns.

To materialize this shift, we propose \sysname, a memory-efficient inference system tailored for long-context dLLMs. \sysname shifts from local, static management to a global, dynamic paradigm. We introduce a mask-only logits kernel that eliminates redundancy by computing logits solely for masked tokens. To handle dynamic bottlenecks, a lazy chunking optimizer employs an online search to split memory-intensive operators only when necessary, minimizing latency overhead. Underpinning these components is a global memory manager utilizing a graph registrar to capture the full computation lifecycle, enabling a unified virtual memory mapping that eliminates external fragmentation.

Our contributions are summarized as follows:

\begin{itemize}
\item We provide the first comprehensive characterization of dLLM memory usage, identifying the bottleneck shift to transient activations and the phenomenon of dynamic memory peaks.
\item We propose \sysname, an inference system that integrates mask-only computation, a lazy chunking strategy, and a global memory manager utilizing a graph registrar to capture the full computation lifecycle for efficient long-context dLLM inference.
\item Extensive evaluations demonstrate that \sysname achieves an average 2.71$\times$ reduction in memory peak-to-average ratio and extends the maximum supportable sequence length by 15.30--32.34$\times$. Crucially, \sysname is training-free and preserves exact model outputs, while simultaneously reducing end-to-end latency by 2.5\%--55.4\%.
\end{itemize}

\textbf{Conflict of Interest Disclosure.} The authors declare that they have no financial conflicts of interest to disclose.

\section{Preliminaries}


This section reviews LLM inference paradigms and memory management techniques, focusing on the fundamental differences between autoregressive (AR) and diffusion-based LLMs (dLLMs) and their distinct memory characteristics.

\subsection{From Autoregression to Diffusion} \label{sec:llm_paradigm}



LLM inference has traditionally followed the autoregressive (AR) paradigm~\cite{yang2025qwen3,dubey2024llama}. Recently, diffusion-based LLMs (dLLMs), such as LLaDA~\cite{nie2025large} and Dream~\cite{ye2025dream}, have emerged with fundamentally different inference workflows. Although both paradigms typically adopt the Transformer architecture~\cite{vaswani2017attention}, they exhibit distinct computation patterns and memory behaviors.





\parab{Autoregression.} AR inference generates tokens sequentially. At each iteration, the model produces a single token, appends it to the sequence, and feeds the extended sequence into the next step. Due to causal dependencies, the effective sequence length grows monotonically until an EOS token or a length limit is reached.

\begin{figure} 
    \centering 
    \includegraphics[width=0.45\textwidth]{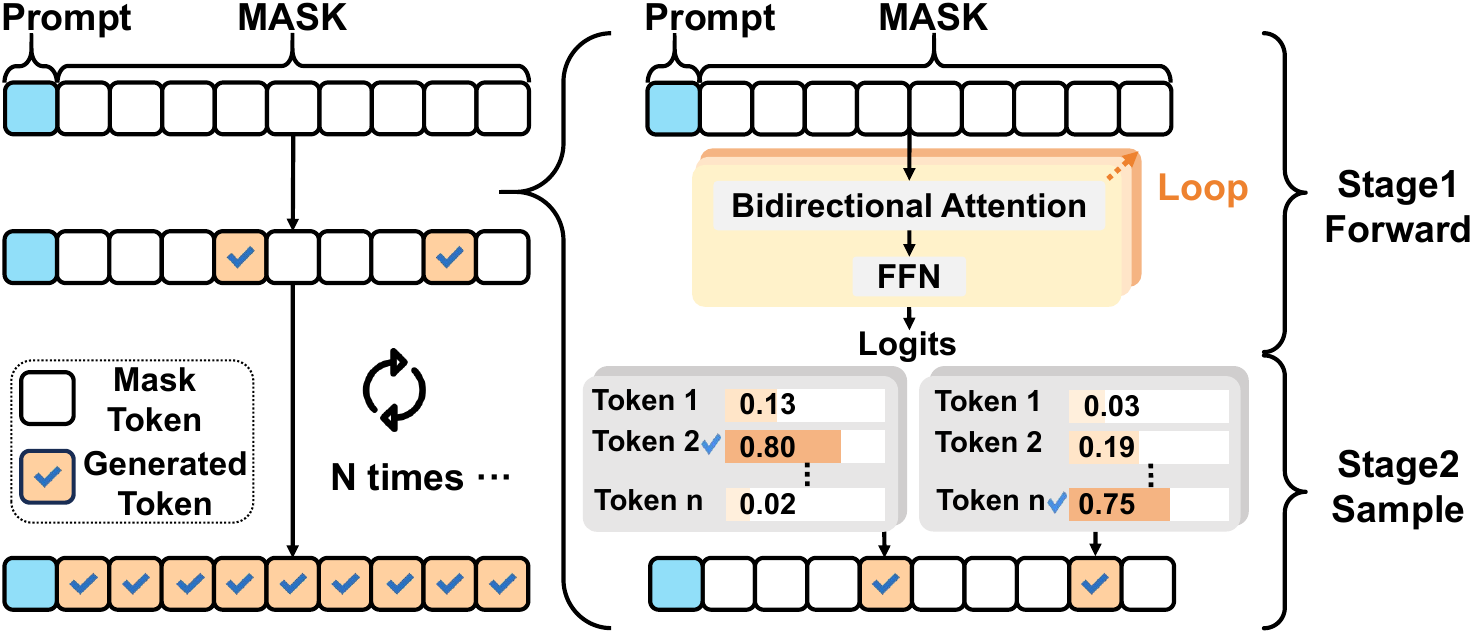} 
    \caption{dLLM inference pipeline.} 
    \label{fig:dllm_inference} 
\end{figure}



\parab{Diffusion.} In contrast, dLLMs perform non-autoregressive, iterative refinement, generating all output tokens jointly. Given a prompt, target length $L$, and inference steps $N$, the process starts from a fully masked sequence of length $L$. Over $N$ steps, masks are progressively replaced with sampled tokens.



As shown in \figref{fig:dllm_inference}, each step consists of two stages: (1) a bidirectional forward pass that produces logits for all masked positions, and (2) a sampling stage that selects and fills a subset of masks. Across iterations, the sequence transitions from fully masked to fully specified.




\parab{Memory Implications.} Both paradigms produce large intermediate tensors, including activations and attention-related states, whose memory footprint often exceeds that of model parameters. Efficient memory management is therefore critical for scalable inference.

\subsection{LLM Memory Management} \label{sec:llm_memory_manage}

LLM inference memory consumption is dominated by two categories: (1) transient activations within a forward pass, and (2) persistent caches such as the KV-Cache in AR decoding. We briefly summarize representative management techniques as follows.



\parab{Transient Activations via Memory Reuse.} Activations are short-lived and only required within a single forward pass. Graph-based memory planning models the computation as a directed acyclic graph (DAG), enabling liveness analysis to identify each tensor's creation and last use. Tensors with non-overlapping lifetimes can safely share memory buffers, reducing peak memory usage. This technique underpins memory optimization in modern inference engines~\cite{ansel2024pytorch,pisarchyk2020efficient,lamprou2023safe}.

\parab{KV-Cache via Paging.} In AR inference, causal attention allows past keys and values to be reused across iterations, forming the KV-Cache~\cite{li2024survey}. Inference consists of a prefill phase that initializes the cache from the prompt and a decode phase that incrementally extends it. Since KV-Cache size grows linearly with sequence length, paging-based methods such as PagedAttention~\cite{kwon2023efficient} store KV tensors in fixed-size blocks, reducing fragmentation and enabling long-context generation.

\section{Observations}
\label{sec:observations}



As discussed in \secref{sec:llm_paradigm}, efficient memory management is critical for long-context inference. However, existing techniques (\secref{sec:llm_memory_manage}), largely designed for autoregressive (AR) models, do not translate to diffusion-based LLMs (dLLMs). Through empirical analysis, we identify four observations that reveal fundamental mismatches between conventional memory management and dLLM inference.



\begin{figure}[!t]
    \centering

    \makebox[0.49\textwidth]{%
    \includegraphics[width=0.36\textwidth]{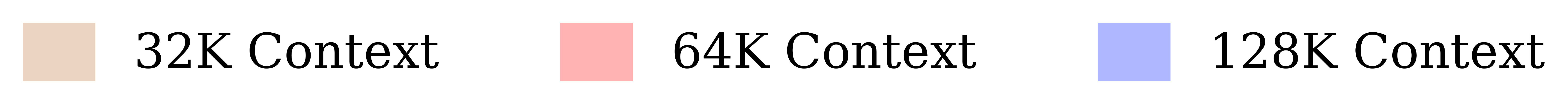} 
    }
    
    \vspace{0.2cm}

    \begin{subfigure}{0.24\textwidth}
        \centering
        \includegraphics[width=\textwidth]{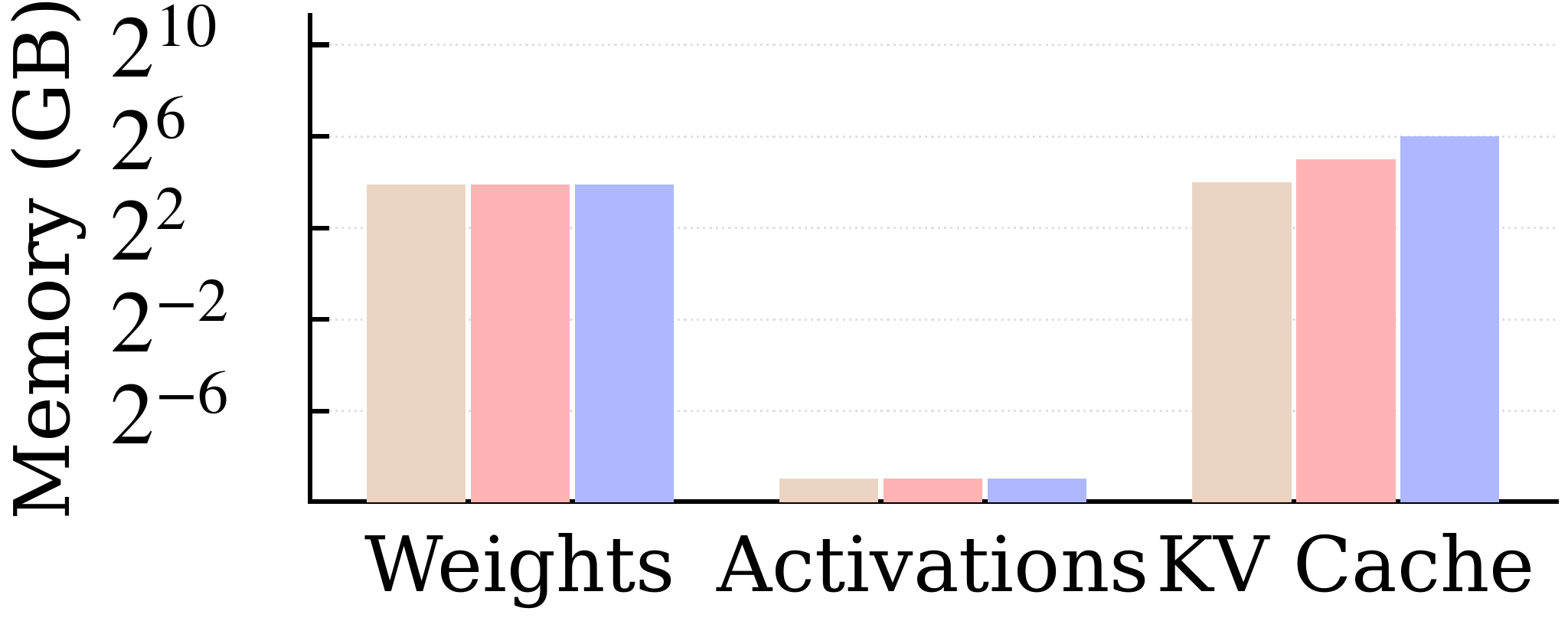}
        \caption{Llama3-8B decoding (AR).}
        \label{fig:arllm_memory}  
    \end{subfigure}%
    \hfill 
    \begin{subfigure}{0.24\textwidth}
        \centering
        \includegraphics[width=\textwidth]{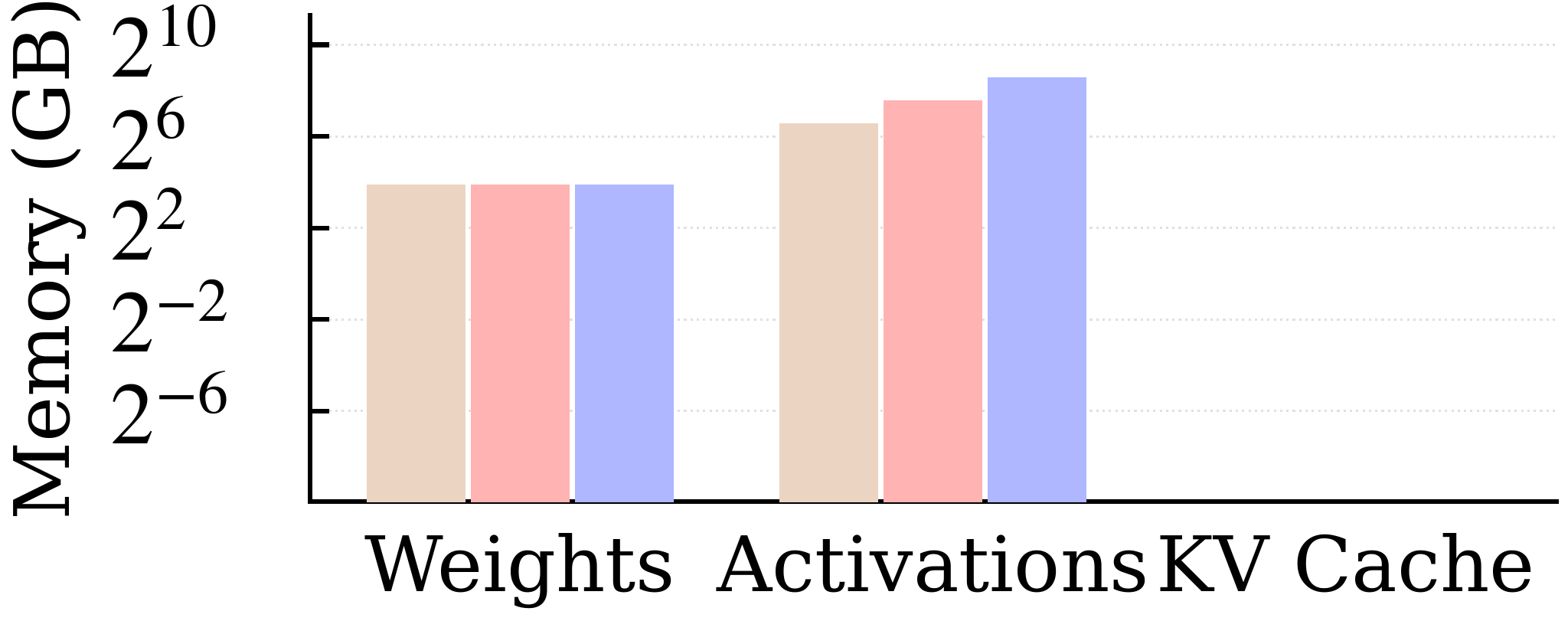}
        \caption{LLaDA-8B (dLLM).}
        \label{fig:dllm_memory}
    \end{subfigure}%

    \caption{Memory breakdown of AR models and dLLMs.}
    \label{fig:memory_breakdown}
\end{figure}

\parab{Observation \#1: The memory bottleneck shifts from KV-Cache to transient activations.} We first compare the memory breakdown of AR and diffusion inference as context length increases. In AR decoding, transient activations are computed only for the newly generated token at each step, while the KV-Cache grows monotonically and dominates memory usage (\figref{fig:arllm_memory}).

In contrast, dLLMs employ bidirectional attention, causing key and value tensors to change across steps. This prevents caching and requires recomputing activations for all tokens at every step. As a result, transient activations dominate memory consumption in dLLMs (\figref{fig:dllm_memory}). This fundamental shift indicates that dLLM inference requires a \textit{transient-activation–centric} memory design rather than the KV-Cache–centric paradigm used in AR systems.



\parab{Observation \#2: Logits computation wastes memory on unmasked tokens.} Given dominance of transient activations, we observe that existing dLLM inference systems still compute logits for all tokens at each step, though only masked tokens are sampled~\cite{nie2025large,ye2025dream,ma2025dinfer}. This eager computation introduces substantial unnecessary memory overhead from unmasked logits.



Analogous to how KV-Cache avoids redundant recomputation in AR models, a natural optimization is \emph{mask-only logits}, which restricts logits computation to masked tokens. However, this is non-trivial in practice, as logits computation requires memory-contiguous inputs~\cite{gale2020sparse}, while masked tokens are typically scattered.


\begin{figure}[t!]
    \centering

    \begin{subfigure}[b]{0.45\textwidth} 
        \centering
        \includegraphics[width=\textwidth]{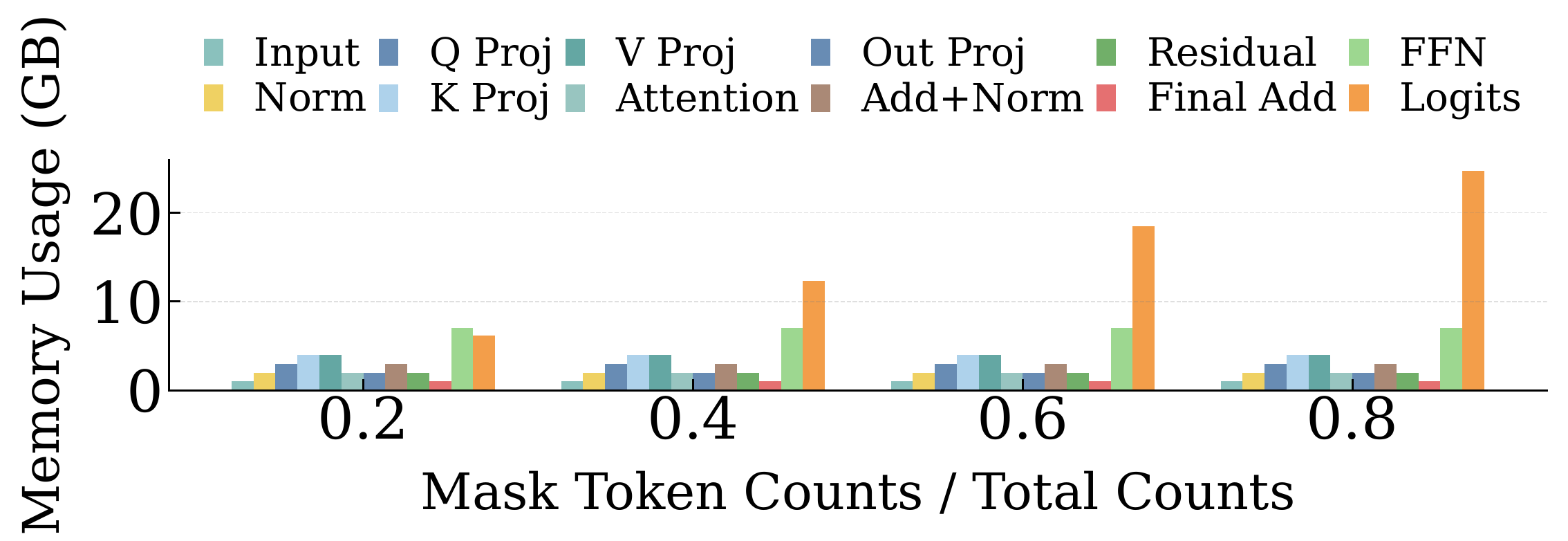}
        \caption{LLaDA-8B.} 
        \label{fig:llada_op}
    \end{subfigure}
    \begin{subfigure}[b]{0.45\textwidth}
        \centering
        \includegraphics[width=\textwidth]{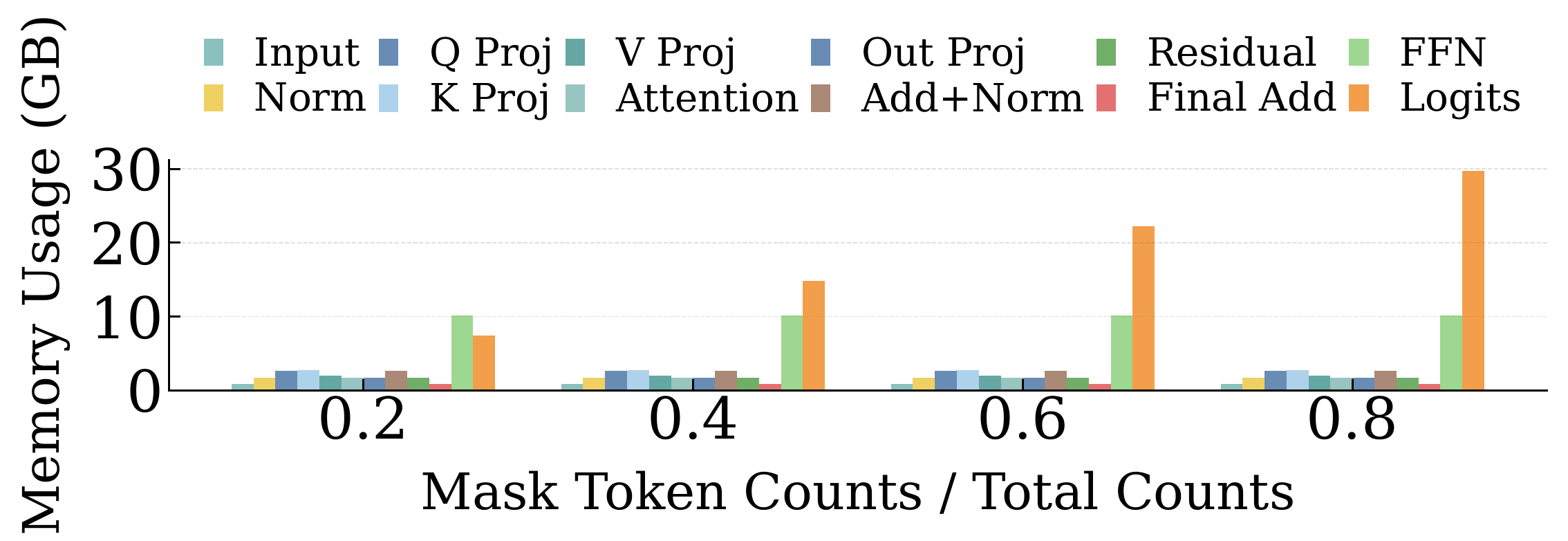}
        \caption{Dream-7B.}
        \label{fig:dream_op}
    \end{subfigure}
      \caption{Dynamic memory bottleneck shift across varying mask ratio $r_m$ at 128k context length.}
    \label{fig:peak_shift}
\end{figure}



\parab{Observation \#3: dLLM inference exhibits dynamic memory peaks.} Even with mask-only logits, inference remains constrained by two transient memory peaks: FFN activations and logits. As shown in \figref{fig:peak_shift}, the dominant peak shifts dynamically with the mask ratio $r_m$ (the fraction of masked tokens). At low $r_m$, FFN activations dominate, whereas at high $r_m$, logits dominate. This behavior is consistent across models such as LLaDA-8B~\cite{nie2025large} and Dream-7B~\cite{ye2025dream}.




This dynamic arises from two factors: (1) large feature dimensions, where both FFN intermediate sizes and logits vocabulary sizes significantly exceed the hidden dimension; and (2) computation asymmetry, where FFNs operate on full sequence while logits are computed only for masked tokens. As $r_m$ varies, the relative contribution of these components shifts, forcing systems to provision memory for worst-case peak and thereby limiting achievable context length.


\begin{figure}[!t]
    \centering
     \begin{minipage}{0.23\textwidth} 
        \centering
        \includegraphics[width=\textwidth]{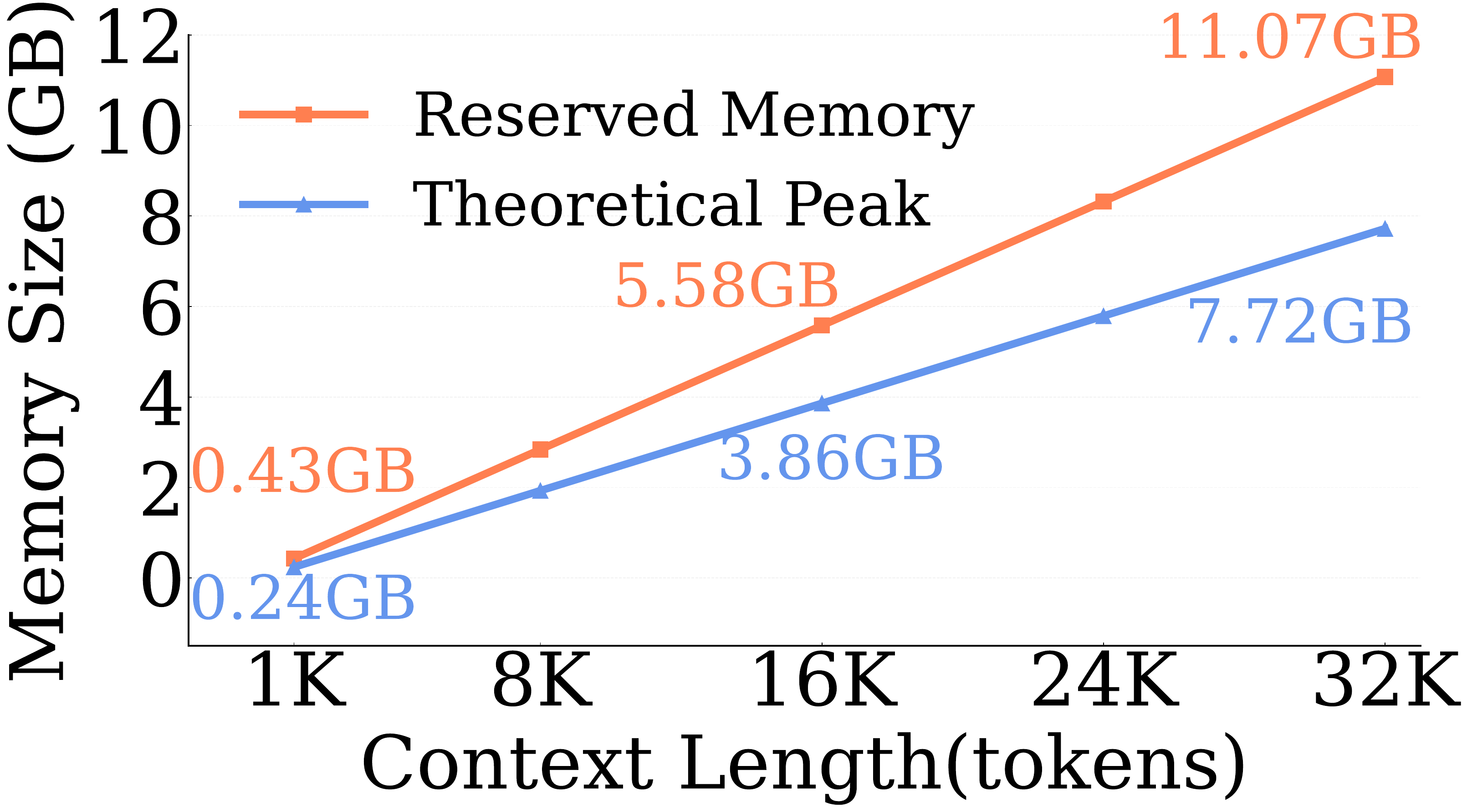} 
         
        \caption{Reserved memory vs. theoretical peak.}
        \label{fig:reserved_vs_used}
    \end{minipage}
    \begin{minipage}{0.23\textwidth} 
        \centering
        \includegraphics[width=\textwidth]{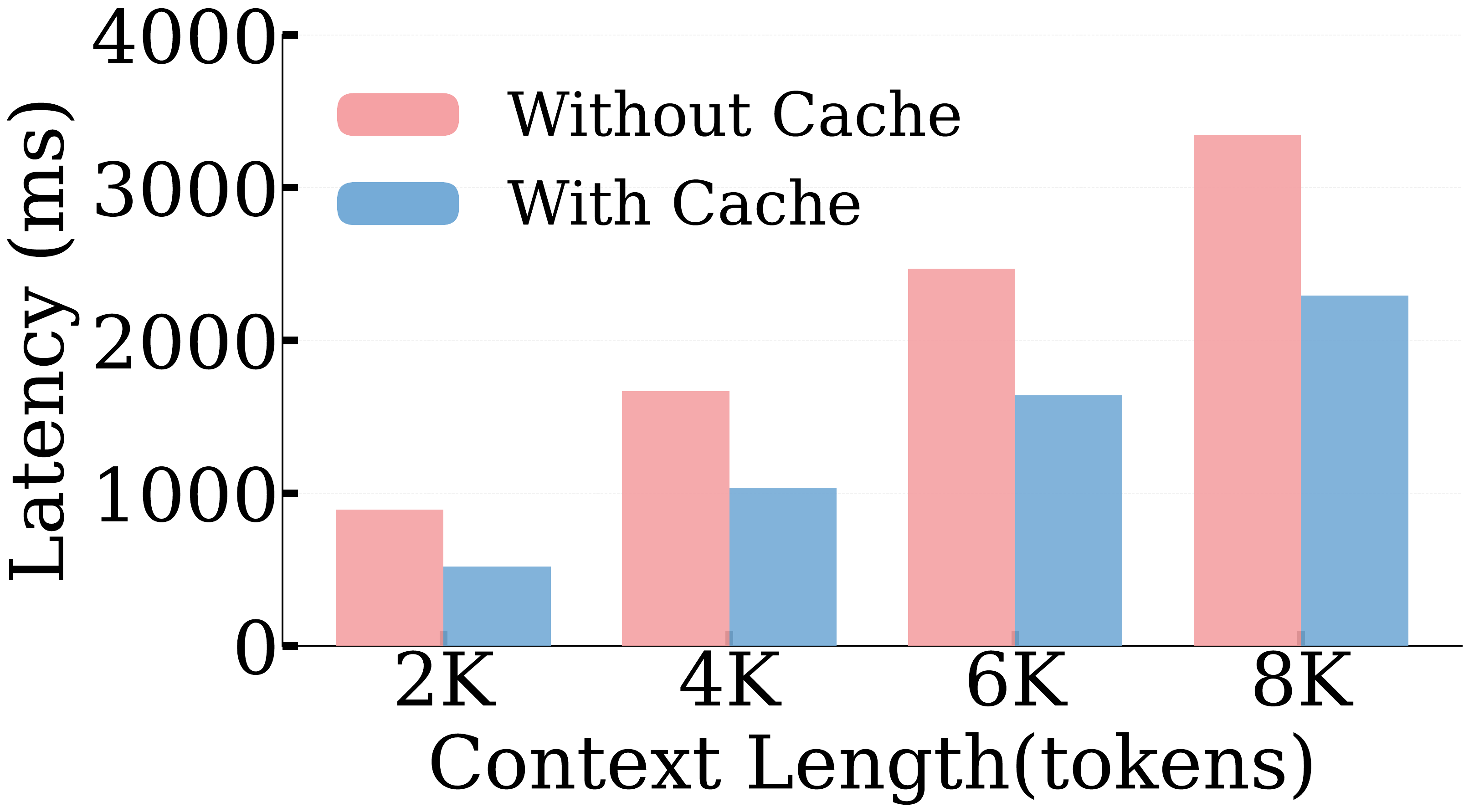} 

        \caption{Latency w/ and w/o segment caching.}
        \label{fig:cache_latency}
    \end{minipage}
\end{figure}




\parab{Observation \#4: Myopic memory planning causes severe fragmentation.} Unlike specialized AR inference engines, dLLM frameworks typically rely on \texttt{torch.compile} and a generic caching allocator~\cite{paszke2019pytorch}. We find that this combination incurs substantial memory overhead. As shown in \figref{fig:reserved_vs_used}, at a 32k context length, reserved memory exceeds the theoretical peak by 43.19\%.

\begin{figure}[!t] 
    \centering 
    \includegraphics[width=0.45\textwidth]{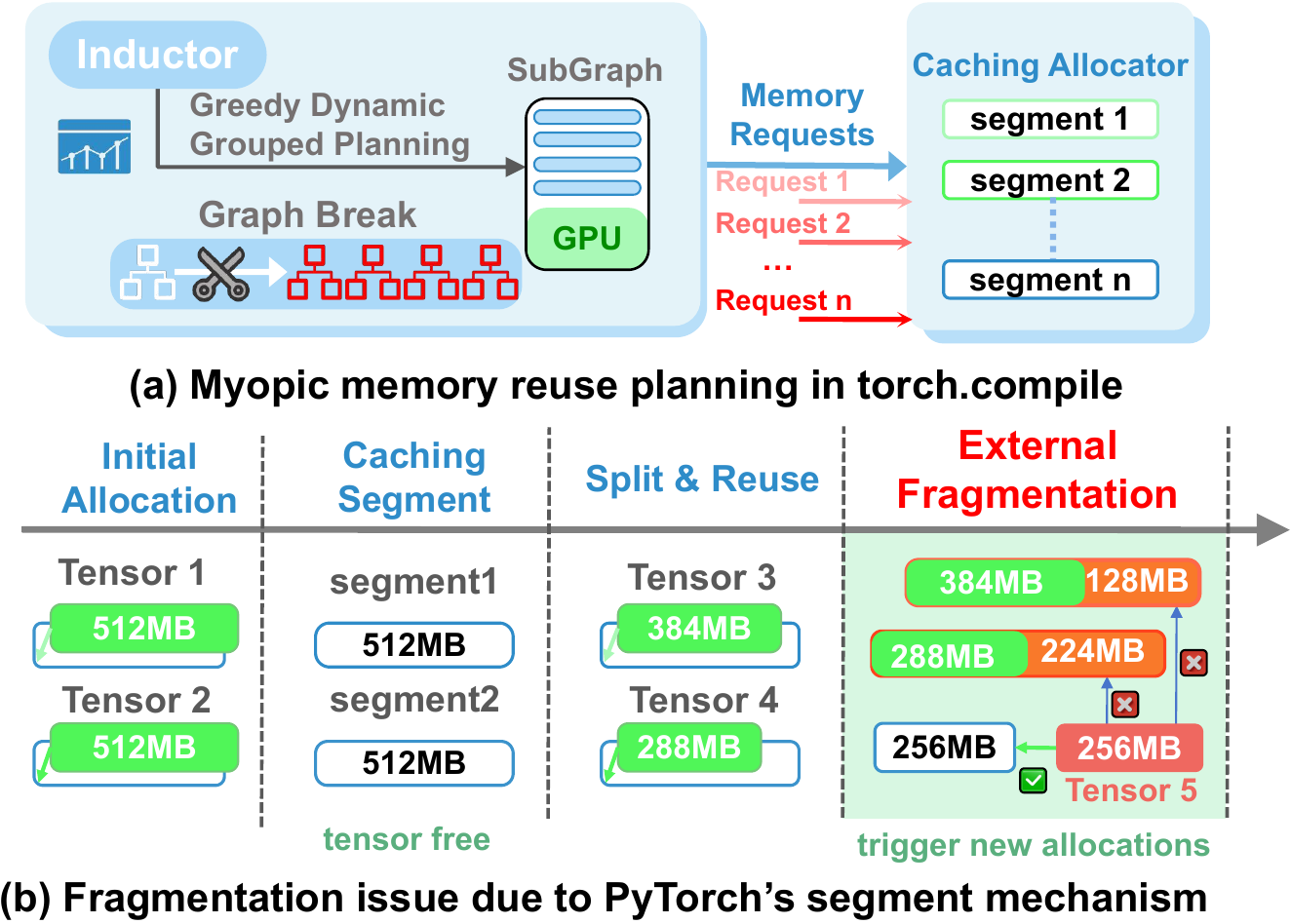} 
    \caption{Illustration of (a) myopic memory reuse planning and (b) the resulting memory fragmentation within the caching allocator.}
    \label{fig:decoupling}
\end{figure}

The root cause is myopic memory planning induced by graph partitioning. Due to greedy grouping and fragile dynamic graph capture~\cite{ansel2024pytorch,liang2025torchtitan}, \texttt{torch.compile} frequently partitions execution into isolated subgraphs (\figref{fig:decoupling}(a)). Each subgraph independently requests contiguous memory, preventing global reuse across the full execution.



This fragmentation propagates to the physical allocator. Dynamic allocation (\eg via \texttt{cudaMalloc}) incurs high runtime overhead, increasing latency by 46\%--72\% (\figref{fig:cache_latency}). Although the allocator caches segments to reduce allocation cost, it preferentially reuses cached segments rather than allocating exact-sized buffers when handling sequential memory requests. This leaves behind non-contiguous fragments (\figref{fig:decoupling}(b)) that cannot satisfy large tensor requests, forcing extra allocations and inflating reserved memory.





\section{Design of \sysname}

\subsection{Overview}

\begin{figure}[!t] 
    \centering 
    \includegraphics[width=0.48\textwidth]{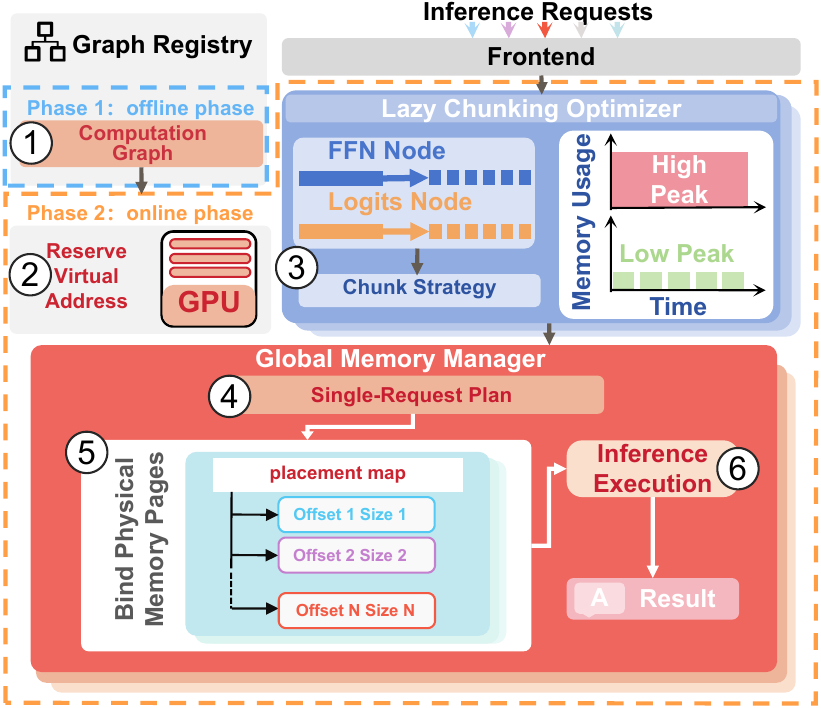} 
    \caption{\sysname overview.} 
    \label{fig:overview} 
\end{figure}


We propose \sysname, a system that enables long-context inference for diffusion-based LLMs. \sysname consists of four tightly integrated components: (1) a \emph{mask-only logits kernel} that computes logits only for masked tokens without auxiliary buffering; (2) a \emph{graph registrar} that explicitly defines a parameterized computation graph to enable whole-graph visibility; (3) a \emph{lazy chunking optimizer} that adaptively chunks memory-intensive operators (FFN and logits) to reduce peak memory with minimal latency overhead; and (4) a \emph{global memory manager} that maps tensors into a virtually contiguous workspace to eliminate fragmentation.


As shown in \figref{fig:overview}, \sysname operates in two phases. In the offline phase, the graph registrar constructs a parameterized graph template using symbolic primitives once per model \circlednum{1}. In the online phase, the global memory manager reserves a contiguous virtual address space \circlednum{2}. Upon receiving an inference request, the lazy chunking optimizer selects a chunking configuration via a bottleneck-driven search \circlednum{3}. The memory manager then derives a global tensor placement map \circlednum{4}, binds physical pages according to the planned peak memory \circlednum{5}, and executes inference \circlednum{6}.





We next describe the four components in detail.

\subsection{Mask-only Logits Kernel}

\begin{figure}[!t] 
    \centering 
    \includegraphics[width=0.48\textwidth]{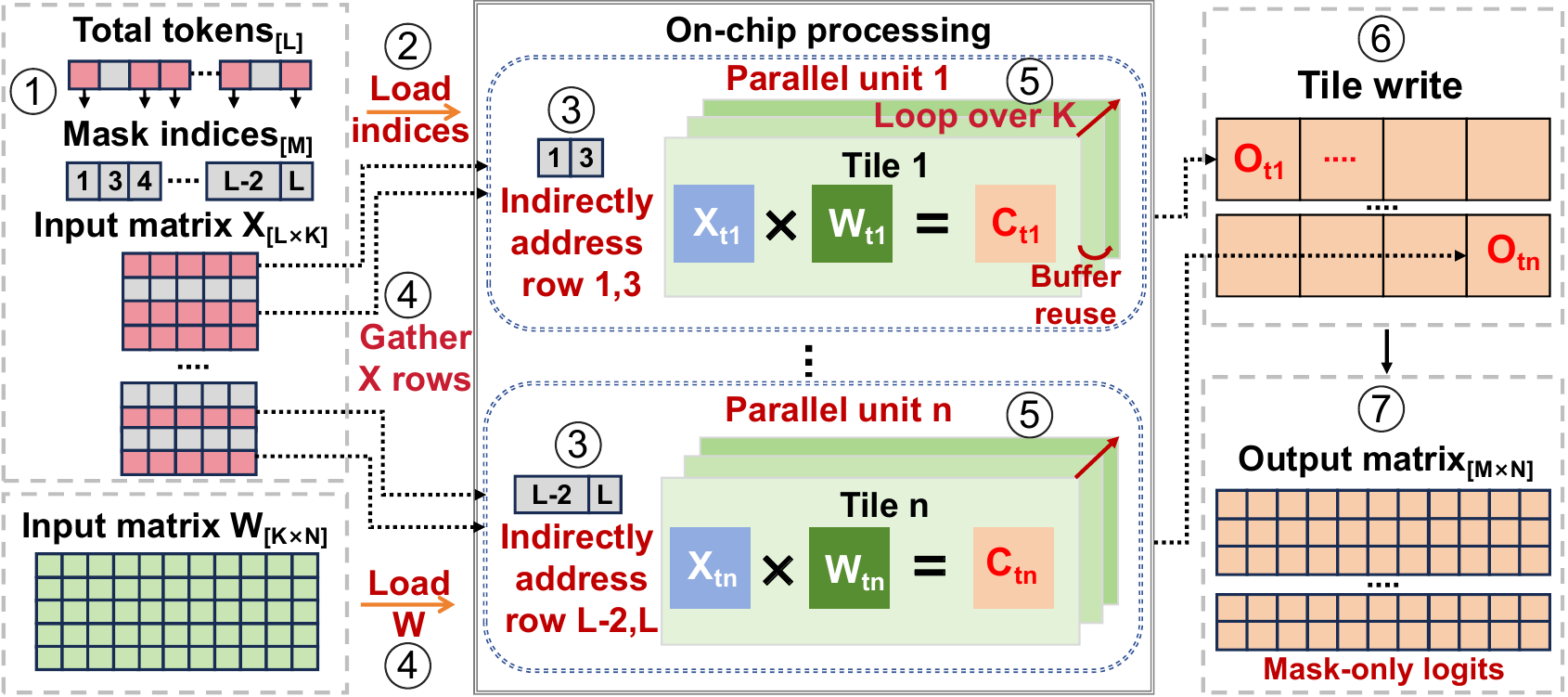} 
    \caption{\sysname's mask-only logits kernel.} 
    \label{fig:mask_only_logits_kernel} 
\end{figure}


To realize mask-only logits (Observation~\#2), \sysname must address a key challenge: logits computation is a GEMM operation that requires memory-contiguous inputs, whereas masked tokens are typically scattered. A naive gather-then-compute approach necessitates a temporary contiguous buffer for the masked hidden states, introducing a memory overhead of $O(N_{\text{mask}} \times D)$, where $N_{\text{mask}}$ is the number of masked tokens and $D$ is the hidden dimension, creating significant redundancy.

\sysname addresses this with a fused gather--GEMM kernel. As illustrated in \figref{fig:mask_only_logits_kernel}, the kernel first extracts mask indices \circlednum{1} and loads them into on-chip memory \circlednum{2}. Using indirect addressing \circlednum{3}, it gathers scattered input rows and fetches weight tiles \circlednum{4} directly into on-chip memory for iterative matrix multiplication \circlednum{5}. The accumulated results are written back directly to the output \circlednum{6}, forming the final logits \circlednum{7} without intermediate buffering. Furthermore, the kernel reduces end-to-end latency by skipping non-masked tokens, with details in \secref{sec:end-to-end-evaluation}.





\subsection{Graph Registrar} \label{sec:graph_registrar}

Supporting lazy chunking and global memory planning requires a complete view of the computation graph. However, compiler-level dynamic capture is often fragile and prone to graph breaks~\cite{ansel2024pytorch,liang2025torchtitan}. \sysname therefore introduces a graph registrar that explicitly defines a parameterized graph template.


\begin{figure}[!t] 
    \centering 
    \includegraphics[width=0.48\textwidth]{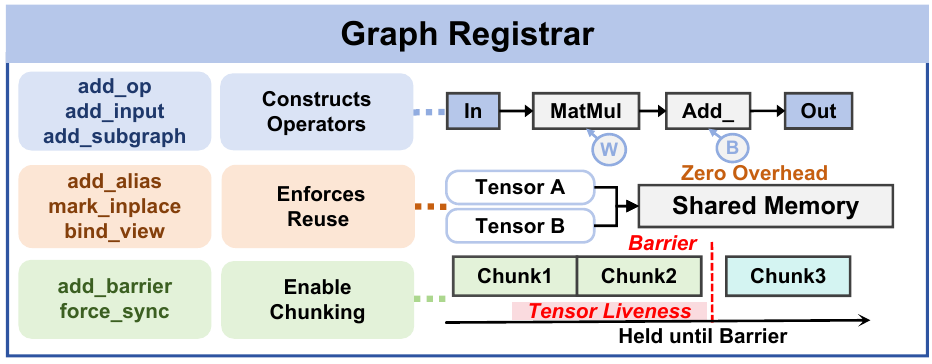} 
    \caption{\sysname's graph registrar.} 
    \label{fig:graph_registrar} 
\end{figure}


As shown in \figref{fig:graph_registrar}, the registrar uses symbolic primitives (\eg \texttt{add\_op}) to construct a template that captures model structure and memory dependencies, while leaving input-dependent dimensions (\eg sequence length) symbolic and instantiated at runtime. This guarantees an unbroken, whole-graph view. The registrar also provides constraint primitives, such as \texttt{add\_alias}, to enforce memory aliasing between operators, and chunking-related primitives to define chunked execution. Control primitives (\eg \texttt{add\_barrier}) extend tensor lifetimes across chunk loops, preventing premature memory reclamation before all chunks complete.

\subsection{Lazy Chunking Optimizer}
\label{sec:lazy_chunking_optimizer}

As shown in \secref{sec:observations}, the memory bottleneck in dLLM inference shifts dynamically between logits and FFN depending on the mask ratio $r_m$. While chunking these operators along the sequence dimension can reduce peak memory, naive chunking suffers from two issues: (1) Latency overhead, as decomposing a single kernel into multiple launches adds execution cost. (2) Misaligned configuration, where static policies might chunk the wrong component as the bottleneck moves (\eg chunking FFN when logits is the peak).

To address these two challenges, \sysname proceeds in three stages. First, it analyzes the latency impact to identify the opportunity to chunk at long context. Second, guided by this, it introduces a lazy chunking strategy to minimize latency overhead. Third, to materialize this strategy under dynamic bottlenecks, we devise an online bottleneck-driven search to find the minimal sufficient chunk configuration.



\begin{figure}
    \centering
    \begin{subfigure}[b]{0.225\textwidth}
        \centering
        \includegraphics[width=\linewidth]{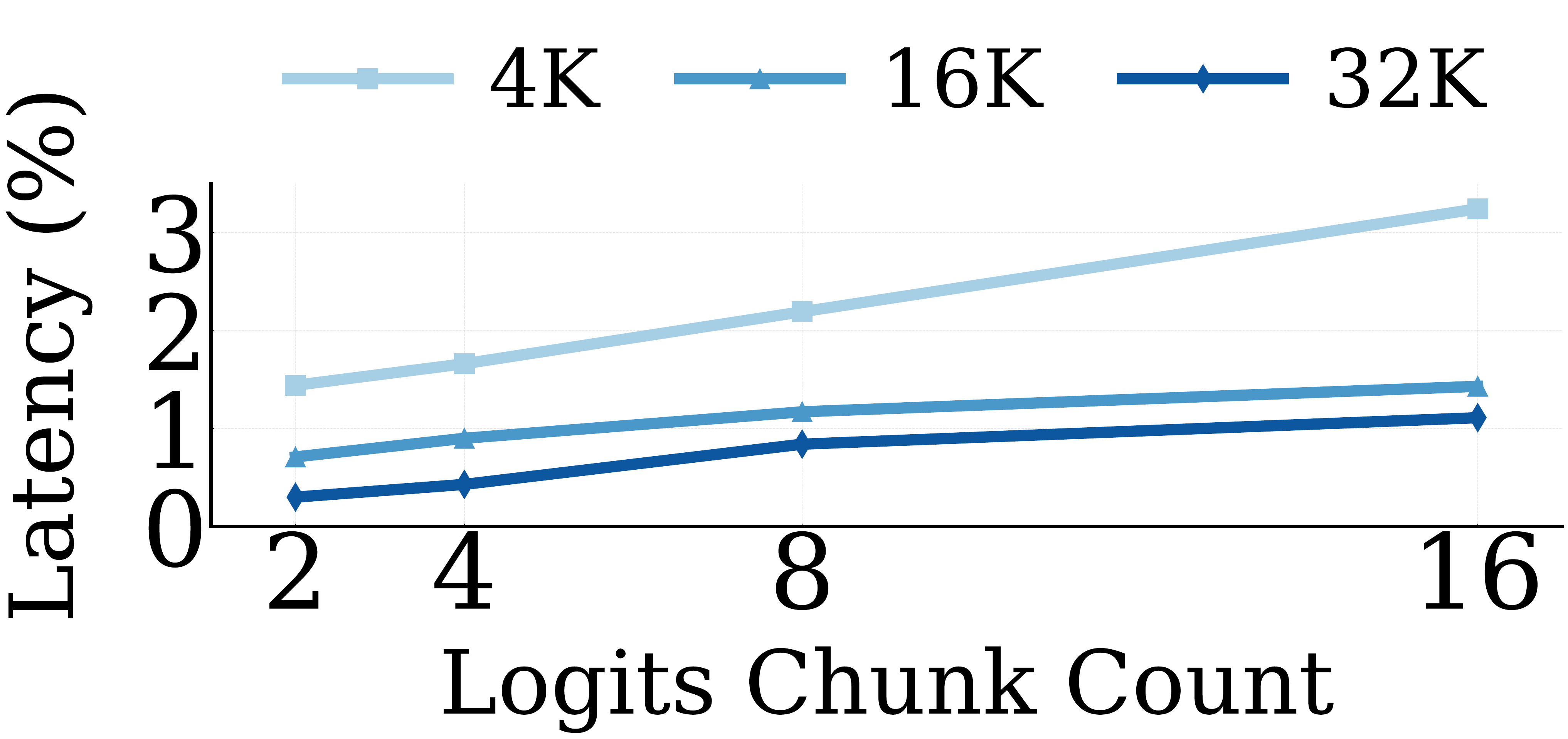} 
        \caption{Impact on logits.}
        \label{fig:logits_latency}
    \end{subfigure}
    \hfill 
    \begin{subfigure}[b]{0.225\textwidth}
        \centering
        \includegraphics[width=\linewidth]{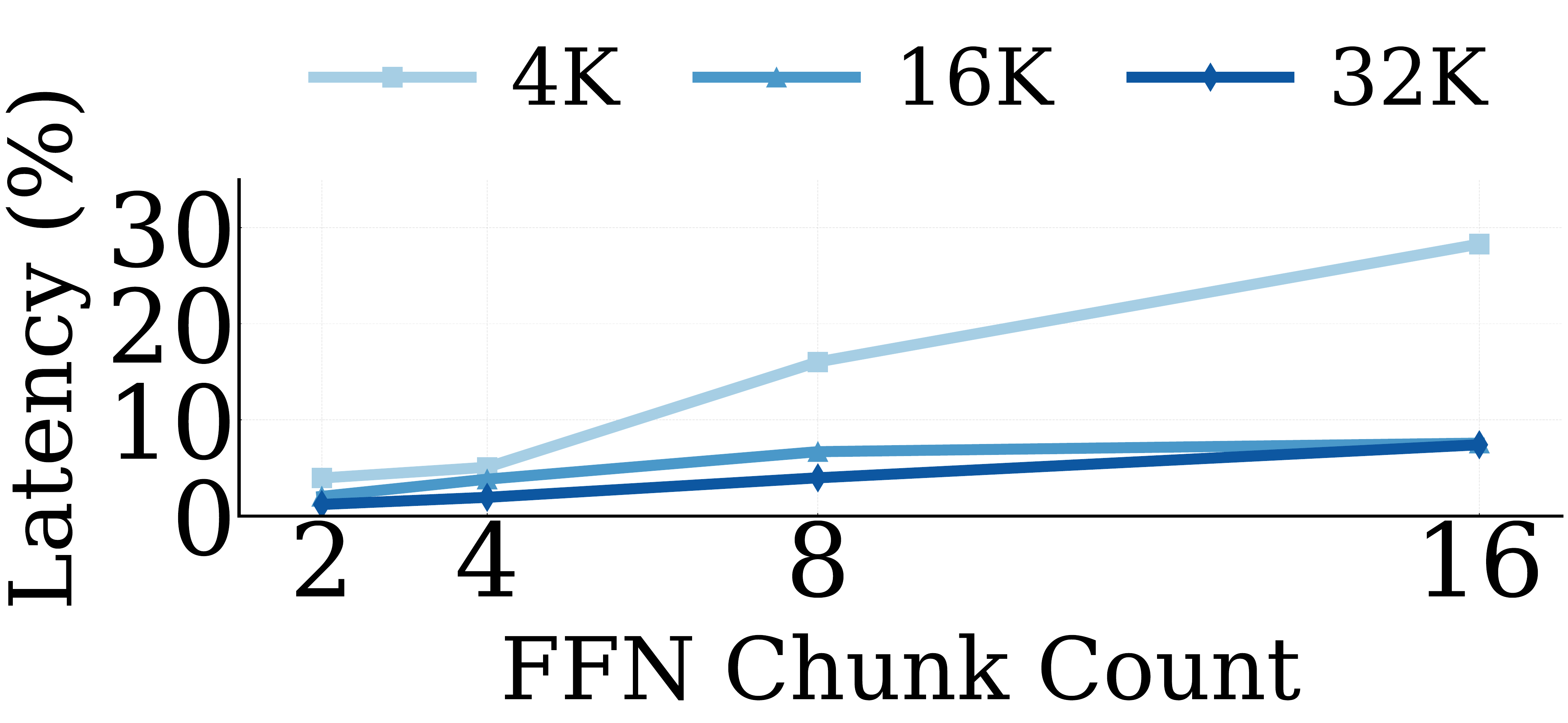} 
        \caption{Impact on FFN.}
        \label{fig:ffn_latency}
    \end{subfigure}
    \caption{Impact of context length and chunk count on the latency of logits and FFN.}
    \label{fig:chunk_latency}
\end{figure}
 

\parab{Opportunity to chunk at long context.} \figref{fig:chunk_latency} reveals that latency increase incurred by chunking diminishes at long contexts. For instance, at 32K context with 4 chunks, the overhead is less than 0.5\% for logits and 2\% for FFNs. This is because at long contexts, the execution time is dominated by massive matrix computations, rendering the fixed kernel launch overhead negligible by comparison.



\parab{Lazy chunking strategy.} Based on this observation, we introduce a lazy chunking strategy. To minimize overhead, we only activate chunking when physical memory is insufficient. Since memory shortages primarily occur at long context, chunking is low-overhead.


When activated, this strategy partitions the logits and FFN computations into smaller segments to reduce peak memory. We denote the number of chunks for these operations as $K_{\text{logits}}$ and $K_{\text{FFN}}$, respectively. To ensure efficiency, we must identify the minimal configuration that exactly supports the target context. However, determining this configuration is non-trivial because the memory bottleneck shifts dynamically with the mask ratio $r_m$. Consequently, we employ an online bottleneck-driven search to solve this dilemma.


\parab{Online bottleneck-driven search.} To instantiate the lazy chunking strategy, we need to find a feasible chunk setting. A naive solution is to brute-force search all combinations. However, this incurs prohibitive overhead. Crucially, we observe that due to memory reuse, the footprint is dictated by the peak and reducing non-peak components yields zero gain. Leveraging this insight, we prune the search space via a bottleneck-driven heuristic. To ensure consistency with runtime execution, we simulate exact buffer sizes by invoking the global memory planner on the registrar's graph.

\begin{algorithm2e}[tb]
\small
\caption{Online Bottleneck-Driven Search}
\label{alg:bottleneck_search}

\SetKwInput{KwInput}{Input}
\SetKwInput{KwOutput}{Output}

\KwInput{Sequence length $T$, mask number $N_{\text{mask}}$, memory capacity $M_{\text{cap}}$}
\KwOutput{Chunk configuration $(K_{\text{logits}}, K_{\text{FFN}})$ or failure}

\BlankLine

Initialize $K_{\text{logits}} \gets 1, K_{\text{FFN}} \gets 1$\;
$P \gets \textsc{SimulatePeak}(T, N_{\text{mask}}, K_{\text{logits}}, K_{\text{FFN}})$\;

\BlankLine

\While{$P > M_{\text{cap}}$}{
    \tcp{Identify component causing the peak}
    Identify component $C_{\text{peak}}$ causing the current peak $P$\;
    
    \tcp{Check if the peak component supports chunking}
    \If{$C_{\text{peak}} \notin \{\textsc{Logits}, \textsc{FFN}\}$}{
        \tcp{Insufficient memory capacity}
        \Return failure; 
    }
    
    \tcp{Adjust configuration based on bottleneck}
    \If{$C_{\text{peak}} = \textsc{Logits}$}{
        $K_{\text{logits}} \gets K_{\text{logits}} + 1$\;
    }
    \ElseIf{$C_{\text{peak}} = \textsc{FFN}$}{
        $K_{\text{FFN}} \gets K_{\text{FFN}} + 1$\;
    }
    
    $P \gets \textsc{SimulatePeak}(T, N_{\text{mask}}, K_{\text{logits}}, K_{\text{FFN}})$\;
}

\tcp{Return valid configuration}
\Return $(K_{\text{logits}}, K_{\text{FFN}})$\;

\end{algorithm2e}




As detailed in Algorithm~\ref{alg:bottleneck_search}, the search takes the number of masks and total tokens as inputs. The search iteratively identifies the current peak component, increases its chunk count, and re-evaluates memory usage using the global memory planner. The process terminates once the peak fits within available memory or no further chunking is possible. As validated in \secref{sec:ablation_search}, \sysname with this heuristic matches the maximum supportable context length of \sysname with brute-force search, while the search latency of this heuristic is only 0.28\%--0.32\% of the brute-force search. The derived $K_{\text{logits}}$ and $K_{\text{FFN}}$ are applied to runtime execution.




\subsection{Global Memory Manager}
\label{sec:global_memory_manager}

To eliminate the memory fragmentation described in \secref{sec:observations}, we identify the root cause of this fragmentation as issuing multiple memory requests. This mechanism implicitly delegates the mapping of logical requests to physical segments to the underlying allocator. As the allocator lacks a global view, this delegation inevitably leads to fragmentation.


To avoid fragmentation, we issue a one-shot global allocation to explicitly bypass the dynamic segment mapping of the underlying allocator. To enable this one-shot allocation, we treat the workspace as a unified address space and pre-calculate a tensor placement map specifying the offset and size for every tensor to direct the runtime execution.


\sysname implements this via two core components: (1) a single-request planner, which performs whole-graph analysis to derive a tensor placement map; and (2) a VMM-based allocator, which dynamically maps physical pages to the planned addresses.


\parab{Single-request planner.} Constructing a unified placement map requires tracking tensor lifecycles. We achieve this by deriving the computation graph from the registrar's templates with lazy chunking. This visibility enables us to perform reuse planning, determining all tensors' offsets.

While finding the optimal offsets is an NP-hard problem solvable via integer linear programming (ILP)~\cite{steiner2023model}, our evaluation (\secref{sec:ablation_planning}) indicates that ILP solvers incur prohibitive latency. Therefore, we adopt a first-fit heuristic, similar to existing works~\cite{chen2018tvm,lin2020mcunet}, which reduces planning time to 0.1\%-4.3\% of the ILP while achieving the same supportable context length.



\parab{VMM-based allocator.} Since the planner maps all tensors into a unified address space, the workspace buffer must be contiguous to support it. However, a naive static pre-allocation would lead to substantial memory waste when handling short sequences. To resolve this, \sysname leverages a virtual memory management (VMM) allocator.

Specifically, during initialization, we only reserve a large range of contiguous virtual addresses occupying no physical memory. At runtime, once the planned peak size is known, we dynamically bind physical pages to this virtual range using low-level page mapping APIs. This ensures that the execution sees a contiguous buffer while the physical consumption is strictly limited to the actual needs. 

With the physical memory being ready, \sysname executes inference following the tensor placement map. However, standard PyTorch operators employ opaque memory management (\eg implicitly allocating internal buffers and lacking support for explicit output placement) that obstructs the strict execution of our placement map. To resolve this, we replace them with customized placement-aware operators to enforce absolute control over memory allocation. These operators not only write directly to pre-determined memory locations but also eliminate redundant internal buffering. For instance, it places Dream~\cite{ye2025dream}'s logits shift and concatenation with a token-level kernel that writes partial results directly to their final positions, which reduces memory usage.

\section{Evaluation}
\subsection{Experimental Setup}
\parab{Implementation.} We implemented \sysname based on vLLM 0.10.1~\cite{kwon2023efficient} to use its robust serving capability. To enable dLLM inference, we integrated the official $\mathtt{PyTorch}$ model and bypassed vLLM's KV-Cache components, replacing them with dLLM-specific logic. \sysname supports variable-length inference to eliminate padding and implements iteration-level scheduling for efficient execution. \sysname comprises approximately 11k lines of code, consisting of 9.5k lines of Python, 1.3k lines of C++/CUDA, and 0.5k lines of Triton.

\parab{Hardware Setup and Models.} We evaluated \sysname on an NVIDIA RTX 3090 with 24GB VRAM and an NVIDIA A100 with 40GB VRAM. We select three mainstream dLLMs: LLaDA-8B~\cite{nie2025large}, Dream-7B~\cite{ye2025dream}, and LLaDA-MoE~\cite{zhu2025llada}. As \sysname targets low-level memory management, its performance is associated with computation graphs and tensor shapes, remaining agnostic to semantic content. Therefore, we evaluate workloads composed of varying prompt and total lengths to cover diverse memory patterns.



\begin{figure*}[!t] 
    \centering 
     \includegraphics[width=0.95\textwidth]{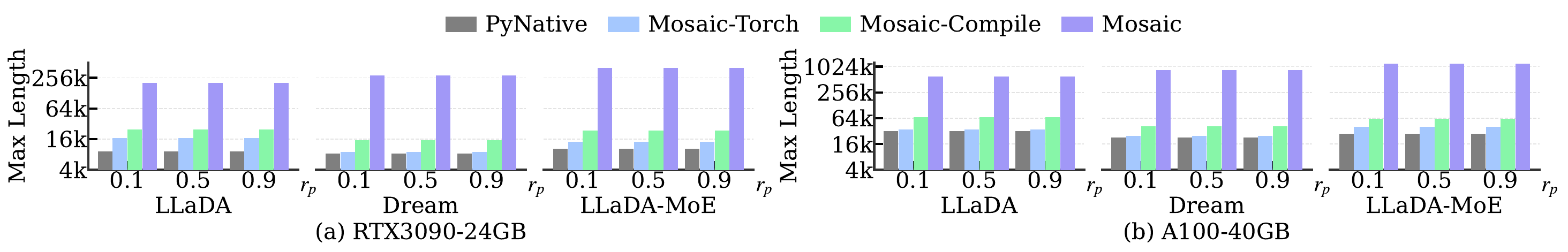}
    \caption{End-to-end performance evaluation comparing $L_{\text{max}}$ of \sysname and baselines across prompt-to-total ratios $r_p$.} 
    \label{fig:e2e} 
\end{figure*}

\parab{Baselines.} We compare \sysname against three baselines: (1) \textsc{PyNative}: The unmodified official \texttt{PyTorch} code for dLLM inference. This measures the raw performance without any serving infrastructure; (2) \sysname-\textsc{Torch}: The \texttt{PyTorch} model code ported onto the vLLM framework to leverage its robust serving infrastructure, but without memory optimizations; (3) \sysname-\textsc{Compile}: This baseline integrates \texttt{torch.compile} into the \sysname-\textsc{Torch}.


\parab{Key Metrics.} We evaluate \sysname and baselines on two metrics: (1) Per-step latency\footnote{Unlike autoregressive models, dLLMs operate via iterative denoising steps without distinct prefill and decode phases, rendering metrics like TTFT and TPOT inapplicable.}, the average time per step; (2) Maximum context length $L_{\text{max}}$, the largest context size supportable before an out-of-memory error.


\subsection{End-to-end Evaluation} \label{sec:end-to-end-evaluation}

For a comprehensive evaluation, we selected requests with varying ratios of prompt length over total length, denoted as $r_p$, for each model. \figref{fig:e2e} illustrates the maximum supportable context length of \sysname and the baselines. As validated in Appendix~\ref{sec:appendix_baselines}, our porting to vLLM incurs no performance penalty; separately, we note that \sysname-\textsc{Compile} extends $L_{\text{max}}$ but suffers from compilation latency. Against this backdrop, \sysname consistently outperforms all alternatives in $L_{\text{max}}$. On average across the three evaluated models and two hardware configurations, its $L_{\text{max}}$ is 32.34, 25.41, and 15.30 times that of \textsc{PyNative}, \sysname-\textsc{Torch}, and \sysname-\textsc{Compile}, respectively. To understand the $L_{\text{max}}$ difference, we conjecture that \sysname's advantage arises from two key factors as follows.



First, existing systems fail to handle memory peaks, thereby bounding the context capacity. In contrast, \sysname employs lazy chunking to mitigate this issue by chunking peak activations when memory is insufficient for longer contexts. As shown in \secref{sec:appendix_par}, we utilize the memory's peak-to-average ratio to quantify the fluctuation of activation memory usage. \sysname exhibits an average 2.71$\times$ reduction in ratio when chunking is enabled; notably, the ratio is lower even before chunking activates because the mask-only kernel reduces logits memory. This effectively resolves spikes, which directly translates to a higher $L_{\text{max}}$.


\begin{figure}[!t]
    \centering
    \includegraphics[width=0.48\textwidth]{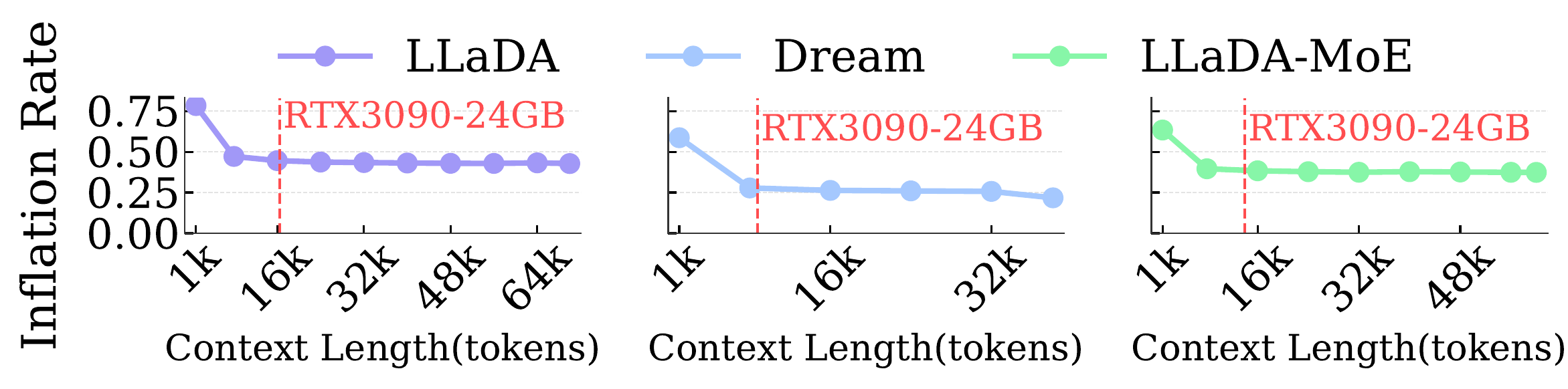}
    \caption{Inflation rates of baselines.}
    \label{fig:inflation_rate}
\end{figure}

Second, existing systems rely on myopic memory planning, leading to external fragmentation and inflated reserved memory compared to the theoretical peak, evidenced by a 21.78\%--78.33\% inflation rate in \figref{fig:inflation_rate}. In contrast, \sysname employs a global manager to eliminate external fragmentation and optimize operators for placement-aware execution (\eg replacing Dream's unique logits shift with a memory-saving, token-level alternative), collectively enabling a larger $L_{\text{max}}$.

Besides, \sysname incurs no compromise on latency. It achieves a 2.5\%--24.9\% reduction compared to \sysname-\textsc{Torch} and \sysname-\textsc{Compile}, attributed to the mask-only logits kernel eliminating redundant computations and the graph registration system bypassing the dynamic graph capture overhead inherent in \texttt{torch.compile}. Furthermore, \sysname reduces latency by 5.9\%--55.4\% compared to \textsc{PyNative}; this significant gain stems from the aforementioned optimizations combined with the inherent engineering efficiencies of the vLLM framework.


\subsection{Ablation Study}

We ablate \sysname’s components to quantify their contributions to maximum context length $L_{\text{max}}$ and latency. Due to space limits, we report results for the prompt-to-total ratio $r_p{=}0.5$; other $r_p$ show similar trends.



\begin{figure}[!t]
    \centering
    \includegraphics[width=0.48\textwidth]{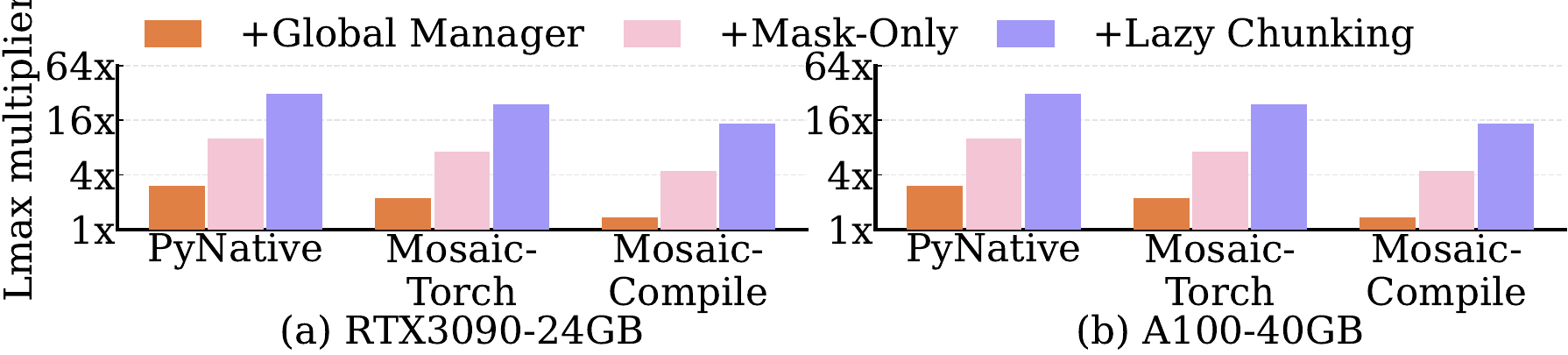}
    \caption{Impact of the global memory manager, mask-only kernel, and lazy chunking on $L_{\text{max}}$.}
    \label{fig:benefits_breakdown}
\end{figure}


\parab{Breakdown of $L_{\text{max}}$ improvements.} To quantify each component's contribution to $L_{\text{max}}$, we performed a cumulative breakdown of the performance gains. For each baseline, we averaged improvements across models and hardware. The reported ranges span the results across the three baselines by incrementally integrating the global memory manager, mask-only kernel, and lazy chunking optimizer.

As shown in \figref{fig:benefits_breakdown}, the global manager increases $L_{\text{max}}$ to 1.41$\times$--3.00$\times$ that of the baselines through global memory reuse coupled with optimized placement-aware operators. Incorporating the mask-only kernel further elevates the capacity to 3.68$\times$--8.04$\times$ relative to the baselines by eliminating redundant logits memory for unmasked tokens, while lazy chunking ultimately boosts $L_{\text{max}}$ to 15.30$\times$--32.34$\times$ over the baselines by mitigating memory spikes.


\begin{figure}[!t]
    \centering
    \includegraphics[width=0.48\textwidth]{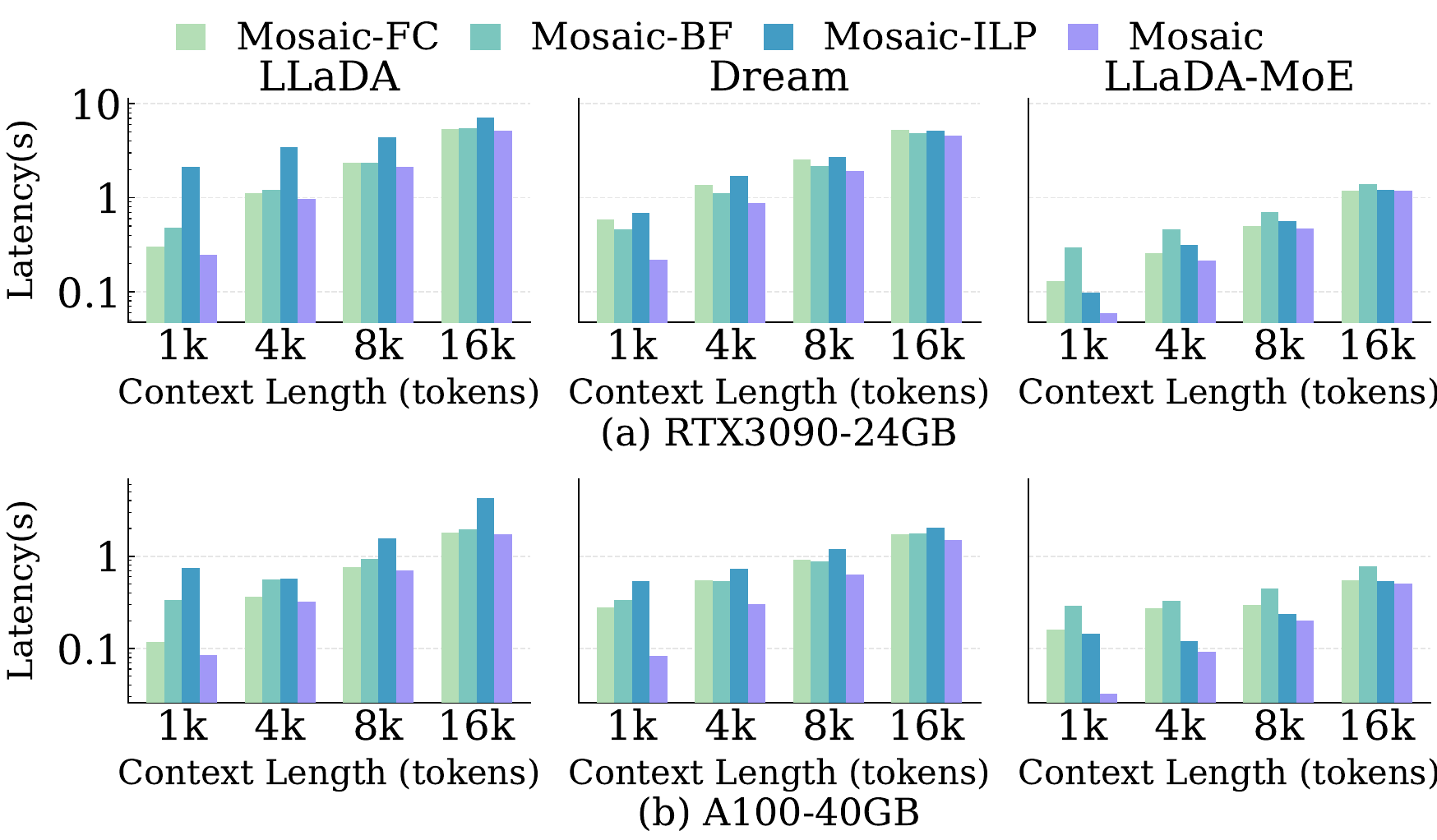}
    \caption{Impact of search, planning, and chunking strategies on end-to-end latency.}
    \label{fig:ablation_latency}
\end{figure}


\parab{Impact of chunking strategy.} \sysname employs a lazy chunking strategy to mitigate the latency overhead induced by chunking operations while maintaining the same $L_{\text{max}}$. To evaluate this, we compare \sysname against \sysname-\textsc{FC}, a baseline that statically applies the configuration supporting $L_{\text{max}}$ across all context lengths.


Our evaluation confirms that both strategies achieve identical $L_{\text{max}}$. As shown in \figref{fig:ablation_latency}, \sysname achieves a 5.5\%--21.5\% latency reduction within the 32k context range compared to \sysname-\textsc{FC}. This reduction is particularly pronounced at shorter context lengths, where the relative overhead of chunking is higher. By dynamically disabling chunking for short contexts, \sysname effectively eliminates this unnecessary overhead.



\begin{figure}[!t]
    \centering
    \includegraphics[width=0.48\textwidth]{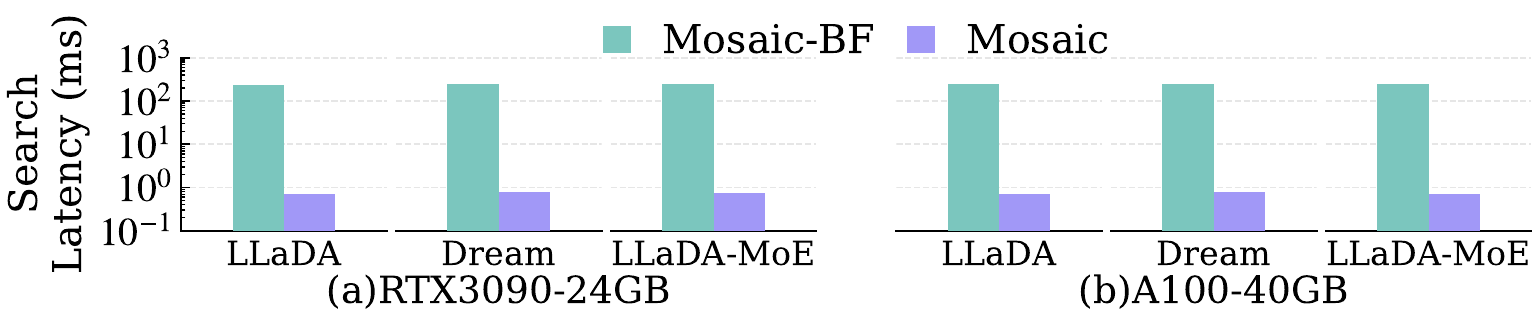}
    \caption{Comparison of search latency.}
    \label{fig:search_latency}
\end{figure}



\parab{Efficiency of bottleneck-driven search.} \label{sec:ablation_search} We compare \sysname’s bottleneck-driven search with \sysname-\textsc{BF}, which uses brute-force search. Both reach the same $L_{\text{max}}$, while \sysname’s search latency is only 0.28\%--0.32\% of brute-force (\figref{fig:search_latency}). This reduces short-context latency by 22.3\%--70.7\% relative to \sysname-\textsc{BF} (\figref{fig:ablation_latency}); the gap narrows at long contexts where computation dominates.

\begin{figure}[!t]
    \centering
    \includegraphics[width=0.48\textwidth]{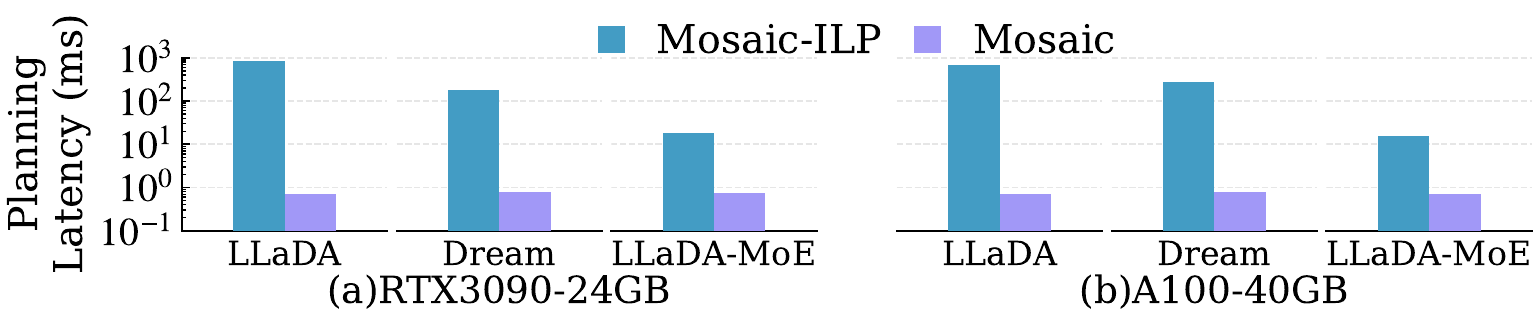}
    \caption{Comparison of planning latency.}
    \label{fig:planning_latency}
\end{figure}



\parab{Efficiency of first-fit planning.} \label{sec:ablation_planning}
\sysname employs a first-fit strategy to determine tensor offsets. We compare it against \sysname-\textsc{ILP}, a baseline utilizing integer linear programming (ILP) for theoretically optimal planning. Our evaluation confirms that both achieve the same $L_{\text{max}}$, while first-fit reduces planning latency to 0.1\%--4.3\% of ILP (\figref{fig:planning_latency}). This translates to a 7.1\%--36.7\% reduction in overall latency versus \sysname-\textsc{ILP} (\figref{fig:ablation_latency}).

\subsection{Scalability and Compatibility}
Having demonstrated the effectiveness of \sysname's core components, we further evaluate its robustness in broader, real-world deployment scenarios. Specifically, to verify \sysname's scalability across hardware configurations and high-concurrency workloads, as well as its compatibility with existing tailored inference stacks, we extend our evaluation to the following three advanced settings.

\parab{Scalability to Multi-GPU Settings.} We extended \sysname to support multi-GPU inference environments. Evaluated on the LLaDA model across 2 GPUs, \sysname achieves a 12.30$\times$ and 14.92$\times$ context length extension under Tensor Parallelism (TP2) and Pipeline Parallelism (PP2), respectively, compared to the \textsc{PyNative} baseline. 

In distributed inference settings, the overall expansion ratio is naturally influenced by the necessary communication buffers. As the context scales to extreme lengths, the memory footprint of these fixed communication buffers (e.g., for inter-GPU tensor passing) grows linearly. While \sysname effectively eliminates the computation activation bottleneck, these unoptimizable communication buffers eventually consume a significant proportion of the GPU memory, inherently bounding the theoretical upper limit of the current context extension. Nevertheless, \sysname still delivers meaningful capacity gains, demonstrating its robustness under distributed environments.

\parab{Compatibility with KV-Cache Frameworks.} While \sysname achieves maximum context extensions of over 32$\times$ on standard lossless baselines where intermediate activations form the strict memory bottleneck, it remains highly complementary to tailored KV-cache stacks. Unlike standard lossless execution, integrating KV caching into dLLMs typically requires lossy approximations. This structurally shifts the memory bottleneck from transient activations to persistent KV states, inherently reducing the optimization scope for activation memory management. 

Nevertheless, by porting \sysname to specialized frameworks, namely Fast-dLLM~\cite{wu2025fast}, dInfer~\cite{ma2025dinfer}, and Elastic-Cache~\cite{nguyen2025attention}, we demonstrate its robust compatibility. Evaluated on an NVIDIA RTX 3090 with 24GB VRAM using the LLaDA-8B model, \sysname still extends the supported context lengths by 3.75$\times$, 3.00$\times$, and 2.27$\times$, respectively. Furthermore, \sysname achieves these capacity extensions under identical hardware configurations without incurring any latency degradation.

\parab{Performance on Batched Workloads.} To validate the practical utility of \sysname under real-world serving scenarios, we further evaluate its scalability with batched workloads. In high-concurrency environments, the memory footprint of intermediate activations scales linearly with the total number of processed tokens. Consequently, under a fixed hardware memory ceiling and a uniform sequence length across all requests, extending the maximum context length mathematically equates to expanding the batch capacity. 

Our empirical evaluations confirm this theoretical equivalence. Specifically, when testing with identically sized requests, \sysname increases the maximum supported batch size by the exact same multipliers observed in the single-request context extensions, achieving an average increase of 32.34$\times$, 25.41$\times$, and 15.30$\times$ over \textsc{PyNative}, \sysname-\textsc{Torch}, and \sysname-\textsc{Compile}, respectively. Furthermore, within the executable range of these baselines, \sysname gracefully handles batched execution without incurring any latency penalty.

\section{Related Work}

\parab{Diffusion Language Models and Optimization.} Text diffusion has evolved from continuous approximations~\cite{li2022diffusion, gong2022diffuseq} to discrete token-space formulations~\cite{austin2021structured, han2023ssd}, culminating in Transformer-based dLLMs~\cite{nie2025large, ye2025dream} that leverage bidirectional attention. While one line of work uses KV-Cache~\cite{wu2025fast} to accelerate inference, it re-introduces the persistent KV bottleneck and may compromise generation quality~\cite{wu2025fast,liu2025dllm,hu2025accelerating}. In contrast, we target the standard bidirectional dLLMs to ensure strict global consistency.


\parab{Activation Memory Reuse.} Mainstream frameworks optimize memory via liveness analysis~\cite{pisarchyk2020efficient, ansel2024pytorch}. 
However, fragile dynamic capture often triggers ``graph breaks''~\cite{ansel2024pytorch} that produce isolated sub-graphs, resulting in suboptimal reuse.



\parab{Persistent State Memory Optimization.} Systems like vLLM~\cite{kwon2023efficient} and vAttention~\cite{prabhu2025vattention} optimize AR inference by managing the persistent KV-Cache via paging or virtual memory. Additionally, KV-Cache offloading alleviates the memory pressure~\cite{hu2025tightllm}, while prefix-aware attention accelerates inference by aggregating KV states~\cite{yi2025pat}. However, these are ill-suited for bidirectional dLLMs.

\section{Conclusion}
In this paper, we identified the fundamental mismatch between conventional memory management and the unique requirements of dLLMs, specifically the shift from KV-Cache to transient activations and dynamic memory peaks. To bridge this gap, we proposed \sysname, a system that harmonizes mask-only computation, global graph planning, and adaptive lazy chunking. It effectively eliminates redundancy and fragmentation, enabling 15.30--32.34$\times$ longer context length and 2.5\%--55.4\% lower latency. Crucially, \sysname is training-free and preserves exact model outputs. Our work paves the way for scalable dLLM inference, making long-context applications more accessible and efficient.


\section*{Acknowledgements}
This work was supported by the National Natural Science Foundation of China under Grant 62572341.

\section*{Impact Statement}

This paper presents work whose goal is to advance the field of Machine
Learning. There are many potential societal consequences of our work, none
which we feel must be specifically highlighted here.

\bibliography{main}

@article{zhao2023survey,
  title={A survey of large language models},
  author={Zhao, Wayne Xin and Zhou, Kun and Li, Junyi and Tang, Tianyi and Wang, Xiaolei and Hou, Yupeng and Min, Yingqian and Zhang, Beichen and Zhang, Junjie and Dong, Zican and others},
  journal={arXiv preprint arXiv:2303.18223},
  volume={1},
  number={2},
  year={2023}
}

@article{minaee2024large,
  title={Large language models: A survey},
  author={Minaee, Shervin and Mikolov, Tomas and Nikzad, Narjes and Chenaghlu, Meysam and Socher, Richard and Amatriain, Xavier and Gao, Jianfeng},
  journal={arXiv preprint arXiv:2402.06196},
  year={2024}
}

@article{nie2025large,
  title={Large language diffusion models},
  author={Nie, Shen and Zhu, Fengqi and You, Zebin and Zhang, Xiaolu and Ou, Jingyang and Hu, Jun and Zhou, Jun and Lin, Yankai and Wen, Ji-Rong and Li, Chongxuan},
  journal={arXiv preprint arXiv:2502.09992},
  year={2025}
}

@article{zhu2025llada,
  title={Llada-moe: A sparse moe diffusion language model},
  author={Zhu, Fengqi and You, Zebin and Xing, Yipeng and Huang, Zenan and Liu, Lin and Zhuang, Yihong and Lu, Guoshan and Wang, Kangyu and Wang, Xudong and Wei, Lanning and others},
  journal={arXiv preprint arXiv:2509.24389},
  year={2025}
}

@article{ye2025dream,
  title={Dream 7b: Diffusion large language models},
  author={Ye, Jiacheng and Xie, Zhihui and Zheng, Lin and Gao, Jiahui and Wu, Zirui and Jiang, Xin and Li, Zhenguo and Kong, Lingpeng},
  journal={arXiv preprint arXiv:2508.15487},
  year={2025}
}

@article{sahoo2024simple,
  title={Simple and effective masked diffusion language models},
  author={Sahoo, Subham and Arriola, Marianne and Schiff, Yair and Gokaslan, Aaron and Marroquin, Edgar and Chiu, Justin and Rush, Alexander and Kuleshov, Volodymyr},
  journal={Advances in Neural Information Processing Systems},
  volume={37},
  pages={130136--130184},
  year={2024}
}

@article{yu2025discrete,
  title={Discrete Diffusion in Large Language and Multimodal Models: A Survey},
  author={Yu, Runpeng and Li, Qi and Wang, Xinchao},
  journal={arXiv preprint arXiv:2506.13759},
  year={2025}
}

@article{liu2025longllada,
  title={Longllada: Unlocking long context capabilities in diffusion llms},
  author={Liu, Xiaoran and Song, Yuerong and Liu, Zhigeng and Huang, Zengfeng and Guo, Qipeng and He, Ziwei and Qiu, Xipeng},
  journal={arXiv preprint arXiv:2506.14429},
  year={2025}
}

@article{he2025ultrallada,
  title={Ultrallada: Scaling the context length to 128k for diffusion large language models},
  author={He, Guangxin and Nie, Shen and Zhu, Fengqi and Zhao, Yuankang and Bai, Tianyi and Yan, Ran and Fu, Jie and Li, Chongxuan and Yuan, Binhang},
  journal={arXiv preprint arXiv:2510.10481},
  year={2025}
}

@article{xie2025dream,
  title={Dream-coder 7b: An open diffusion language model for code},
  author={Xie, Zhihui and Ye, Jiacheng and Zheng, Lin and Gao, Jiahui and Dong, Jingwei and Wu, Zirui and Zhao, Xueliang and Gong, Shansan and Jiang, Xin and Li, Zhenguo and others},
  journal={arXiv preprint arXiv:2509.01142},
  year={2025}
}

@inproceedings{kwon2023efficient,
  title={Efficient memory management for large language model serving with pagedattention},
  author={Kwon, Woosuk and Li, Zhuohan and Zhuang, Siyuan and Sheng, Ying and Zheng, Lianmin and Yu, Cody Hao and Gonzalez, Joseph and Zhang, Hao and Stoica, Ion},
  booktitle={Proceedings of the 29th symposium on operating systems principles},
  pages={611--626},
  year={2023}
}

@inproceedings{prabhu2025vattention,
  title={vattention: Dynamic memory management for serving llms without pagedattention},
  author={Prabhu, Ramya and Nayak, Ajay and Mohan, Jayashree and Ramjee, Ramachandran and Panwar, Ashish},
  booktitle={Proceedings of the 30th ACM International Conference on Architectural Support for Programming Languages and Operating Systems, Volume 1},
  pages={1133--1150},
  year={2025}
}

@article{pisarchyk2020efficient,
  title={Efficient memory management for deep neural net inference},
  author={Pisarchyk, Yury and Lee, Juhyun},
  journal={arXiv preprint arXiv:2001.03288},
  year={2020}
}

@inproceedings{ansel2024pytorch,
  title={Pytorch 2: Faster machine learning through dynamic python bytecode transformation and graph compilation},
  author={Ansel, Jason and Yang, Edward and He, Horace and Gimelshein, Natalia and Jain, Animesh and Voznesensky, Michael and Bao, Bin and Bell, Peter and Berard, David and Burovski, Evgeni and others},
  booktitle={Proceedings of the 29th ACM International Conference on Architectural Support for Programming Languages and Operating Systems, Volume 2},
  pages={929--947},
  year={2024}
}

@article{yang2025qwen3,
  title={Qwen3 technical report},
  author={Yang, An and Li, Anfeng and Yang, Baosong and Zhang, Beichen and Hui, Binyuan and Zheng, Bo and Yu, Bowen and Gao, Chang and Huang, Chengen and Lv, Chenxu and others},
  journal={arXiv preprint arXiv:2505.09388},
  year={2025}
}

@article{dubey2024llama,
  title={The llama 3 herd of models},
  author={Grattafiori, Aaron and Dubey, Abhimanyu and Jauhri, Abhinav and Pandey, Abhinav and Kadian, Abhishek and Al-Dahle, Ahmad and Letman, Aiesha and Mathur, Akhil and Schelten, Alan and Vaughan, Alex and others},
  journal={arXiv preprint arXiv:2407.21783},
  year={2024}
}

@article{lamprou2023safe,
  title={Safe optimized static memory allocation for parallel deep learning},
  author={Lamprou, Ioannis and Zhang, Zhen and de Juan, Javier and Yang, Hang and Lai, Yongqiang and Filhol, Etienne and Bastoul, Cedric},
  journal={Proceedings of Machine Learning and Systems},
  volume={5},
  pages={305--324},
  year={2023}
}

@article{li2024survey,
  title={A survey on large language model acceleration based on kv cache management},
  author={Li, Haoyang and Li, Yiming and Tian, Anxin and Tang, Tianhao and Xu, Zhanchao and Chen, Xuejia and Hu, Nicole and Dong, Wei and Li, Qing and Chen, Lei},
  journal={arXiv preprint arXiv:2412.19442},
  year={2024}
}

@article{ma2025dinfer,
  title={dinfer: An efficient inference framework for diffusion language models},
  author={Ma, Yuxin and Du, Lun and Wei, Lanning and Chen, Kun and Xu, Qian and Wang, Kangyu and Feng, Guofeng and Lu, Guoshan and Liu, Lin and Qi, Xiaojing and others},
  journal={arXiv preprint arXiv:2510.08666},
  year={2025}
}

@article{paszke2019pytorch,
  title={Pytorch: An imperative style, high-performance deep learning library},
  author={Paszke, Adam and Gross, Sam and Massa, Francisco and Lerer, Adam and Bradbury, James and Chanan, Gregory and Killeen, Trevor and Lin, Zeming and Gimelshein, Natalia and Antiga, Luca and others},
  journal={Advances in neural information processing systems},
  volume={32},
  year={2019}
}

@inproceedings{gale2020sparse,
  title={Sparse gpu kernels for deep learning},
  author={Gale, Trevor and Zaharia, Matei and Young, Cliff and Elsen, Erich},
  booktitle={SC20: International Conference for High Performance Computing, Networking, Storage and Analysis},
  pages={1--14},
  year={2020},
  organization={IEEE}
}

@article{li2022diffusion,
  title={Diffusion-lm improves controllable text generation},
  author={Li, Xiang and Thickstun, John and Gulrajani, Ishaan and Liang, Percy S and Hashimoto, Tatsunori B},
  journal={Advances in neural information processing systems},
  volume={35},
  pages={4328--4343},
  year={2022}
}

@article{gong2022diffuseq,
  title={Diffuseq: Sequence to sequence text generation with diffusion models},
  author={Gong, Shansan and Li, Mukai and Feng, Jiangtao and Wu, Zhiyong and Kong, LingPeng},
  journal={arXiv preprint arXiv:2210.08933},
  year={2022}
}

@article{austin2021structured,
  title={Structured denoising diffusion models in discrete state-spaces},
  author={Austin, Jacob and Johnson, Daniel D and Ho, Jonathan and Tarlow, Daniel and Van Den Berg, Rianne},
  journal={Advances in neural information processing systems},
  volume={34},
  pages={17981--17993},
  year={2021}
}

@inproceedings{han2023ssd,
  title={Ssd-lm: Semi-autoregressive simplex-based diffusion language model for text generation and modular control},
  author={Han, Xiaochuang and Kumar, Sachin and Tsvetkov, Yulia},
  booktitle={Proceedings of the 61st Annual Meeting of the Association for Computational Linguistics (Volume 1: Long Papers)},
  pages={11575--11596},
  year={2023}
}

@article{wu2025fast,
  title={Fast-dllm: Training-free acceleration of diffusion llm by enabling kv cache and parallel decoding},
  author={Wu, Chengyue and Zhang, Hao and Xue, Shuchen and Liu, Zhijian and Diao, Shizhe and Zhu, Ligeng and Luo, Ping and Han, Song and Xie, Enze},
  journal={arXiv preprint arXiv:2505.22618},
  year={2025}
}

@article{liu2025dllm,
  title={dllm-cache: Accelerating diffusion large language models with adaptive caching},
  author={Liu, Zhiyuan and Yang, Yicun and Zhang, Yaojie and Chen, Junjie and Zou, Chang and Wei, Qingyuan and Wang, Shaobo and Zhang, Linfeng},
  journal={arXiv preprint arXiv:2506.06295},
  year={2025}
}

@article{hu2025accelerating,
  title={Accelerating diffusion language model inference via efficient kv caching and guided diffusion},
  author={Hu, Zhanqiu and Meng, Jian and Akhauri, Yash and Abdelfattah, Mohamed S and Seo, Jae-sun and Zhang, Zhiru and Gupta, Udit},
  journal={arXiv preprint arXiv:2505.21467},
  year={2025}
}

@article{vaswani2017attention,
  title={Attention is all you need},
  author={Vaswani, Ashish and Shazeer, Noam and Parmar, Niki and Uszkoreit, Jakob and Jones, Llion and Gomez, Aidan N and Kaiser, {\L}ukasz and Polosukhin, Illia},
  journal={Advances in neural information processing systems},
  volume={30},
  year={2017}
}

@inproceedings{steiner2023model,
  title={Model: memory optimizations for deep learning},
  author={Steiner, Benoit and Elhoushi, Mostafa and Kahn, Jacob and Hegarty, James},
  booktitle={International Conference on Machine Learning},
  pages={32618--32632},
  year={2023},
  organization={PMLR}
}

@article{hu2025tightllm,
  title={TightLLM: Maximizing Throughput for LLM Inference via Adaptive Offloading Policy},
  author={Hu, Yitao and Liu, Xiulong and Yang, Guotao and Li, Linxuan and Zeng, Kai and Zhao, Zhixin and Chen, Sheng and Zhao, Laiping and Li, Wenxin and Li, Keqiu},
  journal={IEEE Transactions on Computers},
  year={2025},
  publisher={IEEE}
}

@article{yi2025pat,
  title={PAT: Accelerating LLM Decoding via Prefix-Aware Attention with Resource Efficient Multi-Tile Kernel},
  author={Yi, Jinjun and Zhao, Zhixin and Hu, Yitao and Yan, Ke and Sun, Weiwei and Wang, Hao and Zhao, Laiping and Zhang, Yuhao and Li, Wenxin and Li, Keqiu},
  journal={arXiv preprint arXiv:2511.22333},
  year={2025}
}

@inproceedings{
   liang2025torchtitan,
   title={TorchTitan: One-stop PyTorch native solution for production ready {LLM} pretraining},
   author={Wanchao Liang and Tianyu Liu and Less Wright and Will Constable and Andrew Gu and Chien-Chin Huang and Iris Zhang and Wei Feng and Howard Huang and Junjie Wang and Sanket Purandare and Gokul Nadathur and Stratos Idreos},
   booktitle={The Thirteenth International Conference on Learning Representations},
   year={2025},
   url={https://openreview.net/forum?id=SFN6Wm7YBI}
}

@article{lin2020mcunet,
  title={Mcunet: Tiny deep learning on iot devices},
  author={Lin, Ji and Chen, Wei-Ming and Lin, Yujun and Gan, Chuang and Han, Song and others},
  journal={Advances in neural information processing systems},
  volume={33},
  pages={11711--11722},
  year={2020}
}

@inproceedings{chen2018tvm,
  title={$\{$TVM$\}$: An automated $\{$End-to-End$\}$ optimizing compiler for deep learning},
  author={Chen, Tianqi and Moreau, Thierry and Jiang, Ziheng and Zheng, Lianmin and Yan, Eddie and Shen, Haichen and Cowan, Meghan and Wang, Leyuan and Hu, Yuwei and Ceze, Luis and others},
  booktitle={13th USENIX Symposium on Operating Systems Design and Implementation (OSDI 18)},
  pages={578--594},
  year={2018}
}

@article{nguyen2025attention,
  title={Attention is all you need for kv cache in diffusion llms},
  author={Nguyen-Tri, Quan and Ranjan, Mukul and Shen, Zhiqiang},
  journal={arXiv preprint arXiv:2510.14973},
  year={2025}
}
\bibliographystyle{icml2026}

\newpage
\appendix
\onecolumn 
\section{Appendix}
\subsection{Porting Validation and Latency Analysis}
\label{sec:appendix_baselines}

\begin{figure}[ht]
    \centering
    \includegraphics[width=0.95\textwidth]{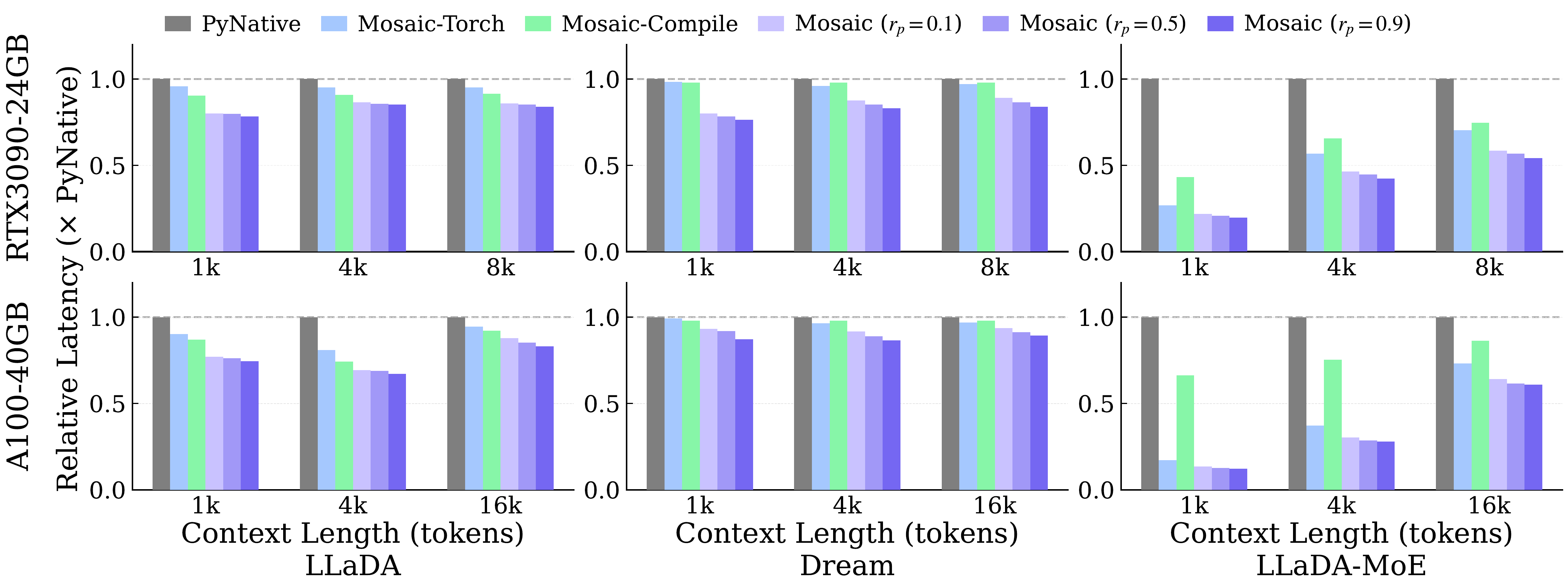}
    \caption{Latency comparison validating that the porting to vLLM incurs no performance penalty, while \sysname-\textsc{Compile} trades latency for increased context length.}
    \label{fig:appendix_baselines}
\end{figure}




\parab{Porting validation.} As shown in \figref{fig:appendix_baselines}, \sysname-\textsc{Torch} significantly outperforms \textsc{PyNative}. Leveraging vLLM's efficient infrastructure such as operator optimizations, it not only reduces latency by 3.1\%--41.2\% but also increases the maximum context length $L_{\text{max}}$ by 7.7\%--48.1\%, as shown in \figref{fig:e2e}. These results confirm that our porting to vLLM incurs no performance penalty and yields comprehensive performance gains.

\parab{Latency analysis.} When further enabling \texttt{torch.compile} on \sysname-\textsc{Torch}, we observe a trade-off. Although \sysname-\textsc{Compile} extends $L_{\text{max}}$ by 45.5\%--97.1\%, it suffers from latency regression due to the overhead associated with dynamic graph capture and compilation, causing its latency to exceed that of \sysname-\textsc{Torch}. In contrast, \sysname incurs no compromise on latency. It achieves a 2.5\%--24.9\% reduction compared to \sysname-\textsc{Torch} and \sysname-\textsc{Compile}, and a 5.9\%--55.4\% reduction compared to \textsc{PyNative}. This significant gain stems from the mask-only logits kernel eliminating redundant computations and the graph registration system bypassing the dynamic graph capture overhead inherent in \texttt{torch.compile}, combined with the inherent engineering efficiencies of the vLLM framework.

\subsection{Peak-to-Average Ratio Analysis} \label{sec:appendix_par}
To further analyze the memory characteristics of dLLMs, we utilize the memory's peak-to-average ratio (PAR) to quantify the fluctuation of activation memory usage. A higher PAR indicates severe memory spikes that limit the maximum supportable context length $L_{\text{max}}$.

\begin{figure}[ht]
    \centering
    \includegraphics[width=0.7\textwidth]{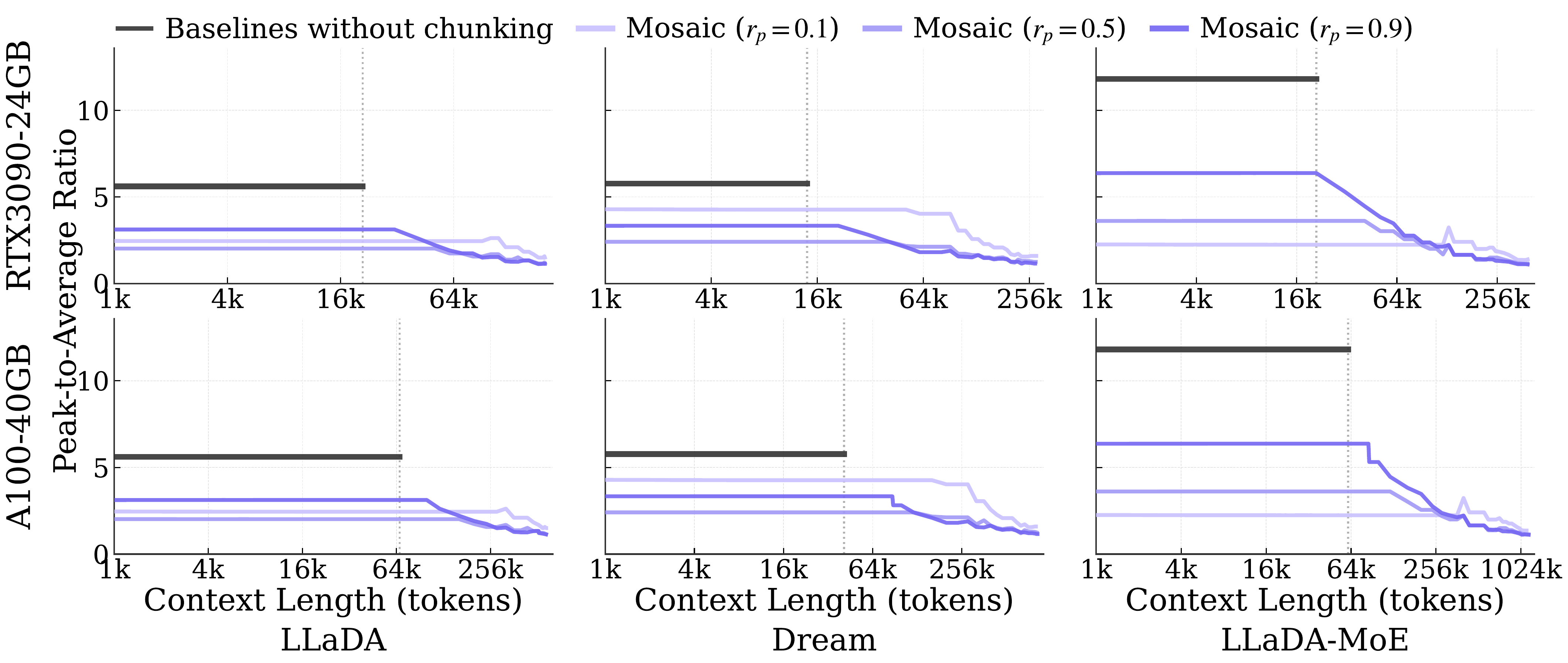}
    \caption{Comparison of activation memory peak-to-average ratio (PAR) across context lengths. Baselines without chunking refer to \textsc{PyNative}, \sysname-\textsc{Torch}, and \sysname-\textsc{Compile}.}
    \label{fig:appendix_par}
\end{figure}

As shown in \figref{fig:appendix_par}, \sysname exhibits a significantly lower ratio when chunking is enabled compared to baselines. Notably, the ratio is lower even before chunking activates because the mask-only kernel reduces logits memory. This reduction in PAR confirms that \sysname effectively smooths out memory spikes, allowing the system to support significantly longer contexts within the same memory budget.


\end{document}